\documentclass[letterpaper]{article} % DO NOT CHANGE THIS
\usepackage{aaai25}  % DO NOT CHANGE THIS
\usepackage{times}  % DO NOT CHANGE THIS
\usepackage{helvet}  % DO NOT CHANGE THIS
\usepackage{courier}  % DO NOT CHANGE THIS
\usepackage[hyphens]{url}  % DO NOT CHANGE THIS
\usepackage{graphicx} % DO NOT CHANGE THIS
\usepackage{booktabs}
\usepackage{multirow}
\usepackage{colortbl}     % For cell colors
\usepackage{arydshln} 
\usepackage{array} 
\usepackage{graphicx}
\usepackage{tcolorbox}
\usepackage{xcolor}
\usepackage{subcaption}
\usepackage[title,toc,page]{appendix}
\usepackage{natbib}  % DO NOT CHANGE THIS AND DO NOT ADD ANY OPTIONS TO IT
\usepackage{caption} % DO NOT CHANGE THIS AND DO NOT ADD ANY OPTIONS TO IT
\DeclareCaptionStyle{ruled}{labelfont=normalfont,labelsep=colon,strut=off} % DO NOT CHANGE THIS
\usepackage{algorithm}
\usepackage{algorithmic}
\usepackage{mdframed}

\usepackage{xcolor}
\newcommand{\answerYes}[1]{\textcolor{blue}{#1}} 
 
\newcommand{\answerNA}[1]{\textcolor{gray}{#1}}

\usepackage{newfloat}
\usepackage{listings}
\floatstyle{ruled}
\newfloat{listing}{tb}{lst}{}
\floatname{listing}{Listing}
\title{LLM-based Semantic Augmentation for\\ Harmful Content Detection}
\author{
    Elyas Meguellati\textsuperscript{\rm 1},
    Assaad Zeghina\textsuperscript{\rm 2},
    Shazia Sadiq\textsuperscript{\rm 1},
    Gianluca Demartini\textsuperscript{\rm 1}
}

\affiliations{
    \textsuperscript{\rm 1}The University of Queensland, Brisbane, Australia\\
    \textsuperscript{\rm 2}Université de Strasbourg, CNRS, ICube, Strasbourg, France\\
    \{m.meguellati, s.sadiq, g.demartini\}@uq.edu.au,
    aozeghina@unistra.fr
}

\begin{document}

\maketitle

\begin{abstract}
Recent advances in large language models (LLMs) have demonstrated strong performance on simple text classification tasks, frequently under zero-shot settings. However, their efficacy declines when tackling complex social media challenges such as propaganda detection, hateful meme classification, and toxicity identification. Much of the existing work has focused on using LLMs to generate synthetic training data, overlooking the potential of LLM-based text preprocessing and semantic augmentation. In this paper, we introduce an approach that prompts LLMs to clean noisy text and provide context-rich explanations, thereby enhancing training sets without substantial increases in data volume.
We systematically evaluate on the SemEval 2024 multi-label Persuasive Meme dataset and further validate on the Google Jigsaw toxic comments and Facebook hateful memes datasets to assess generalizability. Our results reveal that zero-shot LLM classification underperforms on these high-context tasks compared to supervised models. In contrast, integrating LLM-based semantic augmentation yields performance on par with approaches that rely on human-annotated data, at a fraction of the cost. These findings underscore the importance of strategically incorporating LLMs into machine learning (ML) pipeline for social media classification tasks, offering broad implications for combating harmful content online.

\textbf{\textcolor{red}{Disclaimer:}} This paper contains examples of explicit language that may be disturbing to some readers.
\end{abstract}

\section{Introduction}
Large language models (LLMs) have achieved remarkable performance in a broad range of natural language processing (NLP) tasks, showcasing strong zero-shot and few-shot capabilities. Tasks such as sentiment analysis \cite{kuila-sarkar-2024-deciphering} and named entity recognition \cite{xie-etal-2024-self} have particularly benefited from these advancements, highlighting the potential of LLMs to perform well without extensive task-specific training data. However, when applied to more complex tasks, LLMs often struggle to achieve the same level of performance as fine-tuned or task-specific models. Recent studies have highlighted that while zero-shot and few-shot classification using LLMs is effective, these methods frequently underperform as compared to traditional supervised approaches trained on sufficient labeled data \citep{scius-bertrand2024}. For example, fine-tuning smaller models has been shown to outperform zero-shot and few-shots LLMs across a variety of text classification datasets \citep{edwards2024languagemodelstextclassification, hsieh2023distilling}. While some evidence suggests that LLMs can reduce the need for extensive data labeling in resource-limited scenarios \citep{sushil2024comparativestudyzeroshotinference}, their performance declines in tasks requiring nuanced understanding of multimodal inputs, or complex semantic relationships such as propaganda detection \cite{jose2024large} where \textit{the complexity} lies in capturing linguistic nuances, context-dependencies
and latent meaning, due to richness of dynamic variants and figurative use of language \citep{joshi-etal-2015-harnessing}.

Existing research has predominantly focused on leveraging LLMs to augment datasets by generating synthetic examples \citep{patel-etal-2024-datadreamer}. While this approach has shown promise in improving model performance in data-scarce settings \citep{veselovsky2023generatingfaithfulsyntheticdata}, it often overlooks the potential of LLMs to enrich data by adding context (explanations and relevant information) rather than quantity. Semantic augmentation, such as generating explanations for existing data samples, offers a novel pathway for enhancing the feature space without substantially increasing dataset size. Moreover, data cleaning, a critical preprocessing step, has largely relied on rule-based approaches, which often fail to handle unpredictable errors and anomalies in real-world datasets \citep{borrohou2023data}. LLMs, with their advanced understanding of content, offer an alternative by addressing data quality issues more flexibly and accurately \citep{long-etal-2024-llms}.

This study proposes a novel LLM-based semantic augmentation framework that leverages LLMs to clean and contextualize data for context-dependent classifications. Unlike conventional data augmentation approaches, which aim to increase the size of training datasets, our method focuses on enriching feature representations by generating relevant information for each sample and refining noisy text. We evaluate this approach using the \textit{SemEval 2024 persuasive meme dataset}, a multi-label classification problem that forms a subtask of the \textit{SemEval Propaganda Detection} challenge \citep{dimitrov-etal-2024-semeval}. To further validate the generalizability of our method, we also apply it to the \textit{Google Jigsaw toxic comments dataset}\footnote{\url{kaggle.com/c/jigsaw-toxic-comment-classification-challenge}} and the \textit{Facebook Hateful Meme Challenge dataset}\footnote{\url{ai.facebook.com/datasets/hateful-memes}} \cite{kiela2020hateful}.

% Our findings reveal that LLMs offer a cost-effective alternative to human annotators for semantic augmentation, boosting classification performance. Although zero/few-shot LLM classification struggled with high-context tasks, combining LLM-based semantic augmentation with a task-specific model achieved more effective outcomes.

This paper contributes to expanding our understanding of how LLMs can be effectively integrated into machine learning (ML) workflows for social media data analysis. By identifying the most suitable roles for LLMs within a given task, particularly in improving context and addressing data quality issues, we provide a scalable solution for tackling complex classification challenges in online content moderation, propaganda detection, and beyond. The following research questions  guide our investigation:

\begin{enumerate}
   \item[RQ1:] How does incorporating LLM-based data cleaning and semantic augmentation strategies impact the overall performance of models in context-rich classification tasks?
   \item[RQ2:] Does the proposed LLM-based semantic augmentation approach generalize effectively across related domains on social media?
   \item[RQ3:] Beyond explanations, which other semantic augmentation strategies can be used effectively in domains where LLMs might censor or omit critical language elements?
   \item[RQ4:] Under which settings do zero-shot LLMs underperform in social media classification?
\end{enumerate}

\section{Related Work}

\subsection{LLM-Based Data Augmentation}
Traditional data augmentation (DA) methods, such as rule-based transformations \citep{feng2021survey} and back-translation \citep{kumar2019paraphrasing,yang2020generative}, have been significantly enhanced by LLMs that generate synthetic data, improving model performance across diverse NLP tasks \citep{zhou2024survey}. LLM-based DA has been effectively applied in domains such as healthcare \citep{wang2023large} and finance \citep{li2023chatgpt}, with studies demonstrating that synthetic data from LLMs can achieve performance comparable to human-annotated data, at significantly lower costs. For instance, generating 3,000 samples using GPT-3 costs only \$14.37 and takes 46 minutes, compared to the substantially higher cost and time required for human annotation \citep{ding2023gpt}. Despite these advances, synthetic data often underperforms compared to human-labeled data, especially within specialized or highly technical domains \citep{li2023synthetic,ding2023gpt}. Recent approaches have therefore focused on advanced prompting strategies, such as chain-of-thought \citep{wei2022chain} and self-consistency prompting \citep{wang2022self}, as well as noise reduction frameworks like SunGen \citep{gao2023sungen}, to address these limitations and enhance synthetic dataset quality. Such methods have enabled LLMs to better capture task-specific nuances, showing improvement in areas such as commonsense reasoning \citep{liu2021generated} and specialized domains like oncology physics \citep{holmes2023evaluating}.

Cross-domain capabilities of LLMs have also been extensively studied recently. \citet{wang2023large} evaluated LLM performance on healthcare natural language understanding tasks, demonstrating GPT-4's superior performance over alternative models in classification and named entity recognition. Similarly, in finance, \citet{li2023chatgpt} found that GPT-4 performs strongly in news classification and sentiment analysis tasks, though specialized domain-specific models maintain advantages in narrower contexts. This observation extends across various other domains, with work by \citet{wan2023gpt} showing effectiveness in information extraction, and \citet{bang2023multitask} confirming high performance in question answering tasks.

\subsection{Persuasion and Propaganda Detection}
Persuasion techniques are integral to the broader framework of propaganda, significantly influencing public opinion through social media and news platforms \citep{brattberg2018russian, golovchenko2020cross}. A range of computational approaches have been developed to identify specific propaganda techniques, notably including the Propaganda Techniques Corpus (PTC) introduced by \citet{da2019fine}, which catalogues eighteen distinct persuasion labels in news articles. Transformer-based models, such as RoBERTa-CRF \citep{jurkiewicz2020applicaai} and Multi-Granularity Networks (MGN) \citep{da2019fine}, have demonstrated strong performance in persuasion detection tasks. Recent studies \citep{jose2024large} suggest that while general-purpose LLMs typically underperform compared to specialized transformer architectures such as RoBERTa-CRF, they still show promising results in detecting specific propaganda categories. Notably, popular LLMs (GPT-3.5, GPT-4, and Claude 3 Opus) outperform the MGN baseline in categories like \textit{name-calling}. Moreover, GPT-3.5 and GPT-4 achieve stronger performance in detecting techniques such as \textit{appeal to fear} and \textit{flag-waving}, though they struggle with more subtle persuasive strategies such as \textit{bandwagon} and \textit{doubt}.

These observations are consistent with broader trends in recent literature on LLM capabilities in complex NLP tasks, highlighting both the strengths and remaining challenges of current systems. The evolution of these detection methods underscores the importance of developing more nuanced and context-aware approaches to effectively address increasingly sophisticated persuasion and propaganda strategies, especially within social media environments.

\section{Datasets}

\subsection{SemEval-2024 Persuasive Memes}
The SemEval-2024 Task 4 dataset is the main dataset for this study as it serves as the basis for all experimental analyses. It is designed for hierarchical multi-label classification of persuasion techniques in memes. The dataset includes 22 distinct persuasion techniques, structured hierarchically to represent the complex relationships between \textit{rhetorical} and \textit{psychological} strategies.

The dataset primarily consists of English-language memes, though it includes a smaller subset in other languages. Each sample contains both textual and visual elements, making it a challenging multimodal dataset. The hierarchical structure adds complexity by requiring models to identify specific techniques and their contextual relationships within the hierarchy. The dataset consists of a training set of 7,000 samples and a test set of 1,500 samples. Hierarchical F1 (\textbf{H-F1}) is the chosen evaluation measure. \textbf{H-F1} differs from the standard F1 score by taking into account the hierarchical relationships among labels \citep{kiritchenko2006learning}. While the standard F1 score evaluates precision and recall for flat, independent labels, \textbf{H-F1} adjusts the evaluation to reflect the hierarchical dependencies between persuasion techniques. This ensures that errors at higher levels in the hierarchy are treated differently from errors at lower levels, providing a more nuanced assessment of the model’s performance.

\subsection{Validation Datasets}
We use additional validation datasets to assess the generalizability of the proposed approach by applying the baselines and best-performing configurations derived from our experiments on the SemEval-2024 Persuasion dataset. They provide a broader evaluation benchmark for the proposed method's effectiveness on downstream tasks across different online contexts.

\subsubsection{Google Jigsaw Toxic Comments}
The Jigsaw Toxic Comments dataset is used to validate the approach in a purely textual domain focused on toxic comment classification. The dataset contains English-language Wikipedia comments, annotated for six categories of toxicity: \textit{toxic, severe toxic, obscene, threat, insult,} and \textit{identity hate}. For this study, only English comments are considered. The dataset consists of a training set of 159,571 samples and a test set of 153,164 samples. The evaluation measures are F1-score and ROC-AUC.

\subsubsection{Facebook Hateful Meme Challenge}
The Facebook Hateful Meme Challenge dataset is used to evaluate methods in a multimodal task involving both text and image content. Each meme is classified as \textit{hateful} or \textit{non-hateful}. This study focuses on memes with English text content to validate the generalizability of the proposed method to multimodal hate speech detection. The dataset consists of a training set of 8,500 samples and a test set of 1,500 samples. The evaluation measures are F1-score and ROC-AUC.

\noindent All three datasets are accompanied by predefined, fixed training and test splits provided by the organizers, with detailed label distributions shown in Tables~\ref{tab:semeval_dist}, \ref{tab:jigsaw_dist}, and \ref{tab:facebook_dist} in the Appendix.
%ensuring consistent evaluation across different approaches without split-dependent variance.

\section{Method}

\subsection{LLM-based Cleaning}
To streamline multimodal processing and reduce computational overhead, we substitute image inputs with textual data produced by captioning models BLIP \citep{li2022blip} and GIT \citep{wang2022git}. This approach is motivated by findings by \citet{nguyen2024improving}, which highlight that converting images into textual representations via captioning models not only simplifies training, but also significantly lowers computational complexity and speeds up inferences.

\begin{figure}[ht]
    \begin{mdframed}[linewidth=1pt] % Frame around the figure
        \centering
        \includegraphics[width=\textwidth]{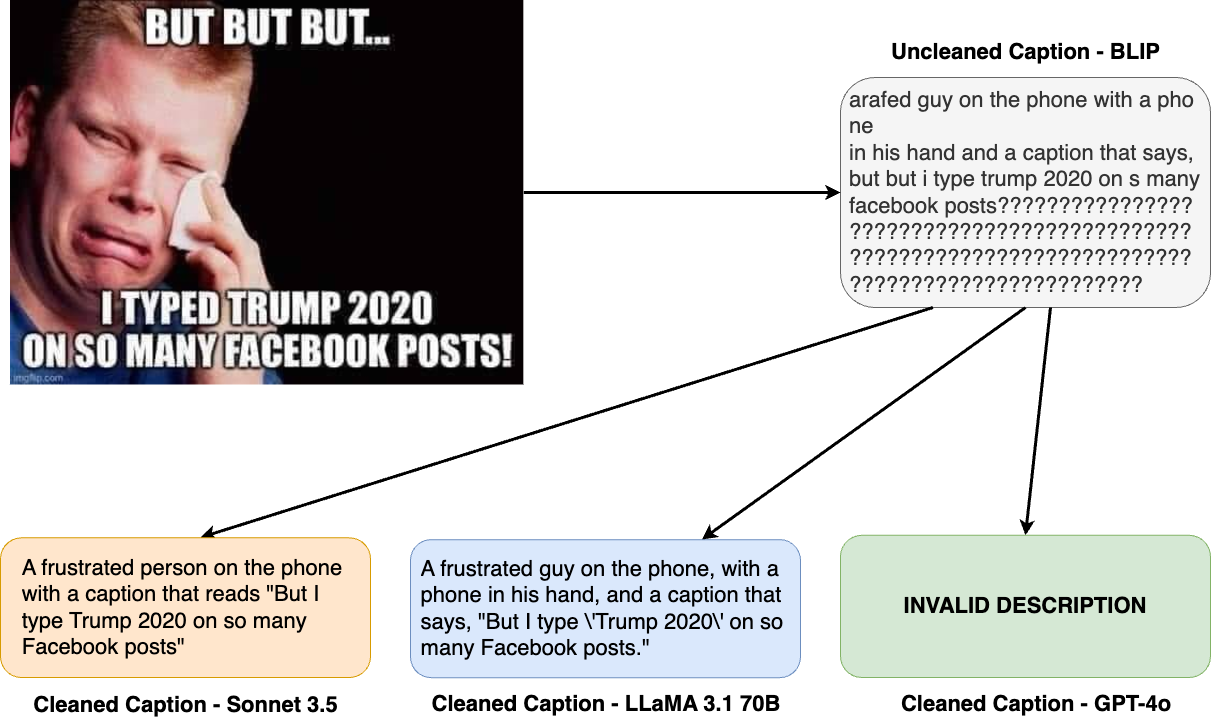} % Full width for the image
    \end{mdframed}
    \caption{LLM-based Cleaning for BLIP-generated Caption.}
    \label{fig:clean_pipeline}
\end{figure}

However, captioning models often produce outputs containing noise and anomalies, such as grammatical errors, redundant phrases, and misaligned descriptions (Figure~\ref{fig:clean_pipeline}). These inconsistencies are challenging to filter out using traditional rule-based methods, thus necessitating a more adaptive approach. Such imperfections make this an ideal scenario for evaluating the benefits of the proposed text cleaning step using LLMs, as the noisy captions provide a realistic testbed for assessing their capability to enhance text quality and ensure task relevance.
The detailed prompt provided for cleaning and correcting captions is shown in Figure~\ref{fig:cleaning_prompt} Appendix.

\begin{figure*}[t]
    \begin{mdframed}[linewidth=1pt] % Frame around the figure
        \centering
        \includegraphics[width=\textwidth]{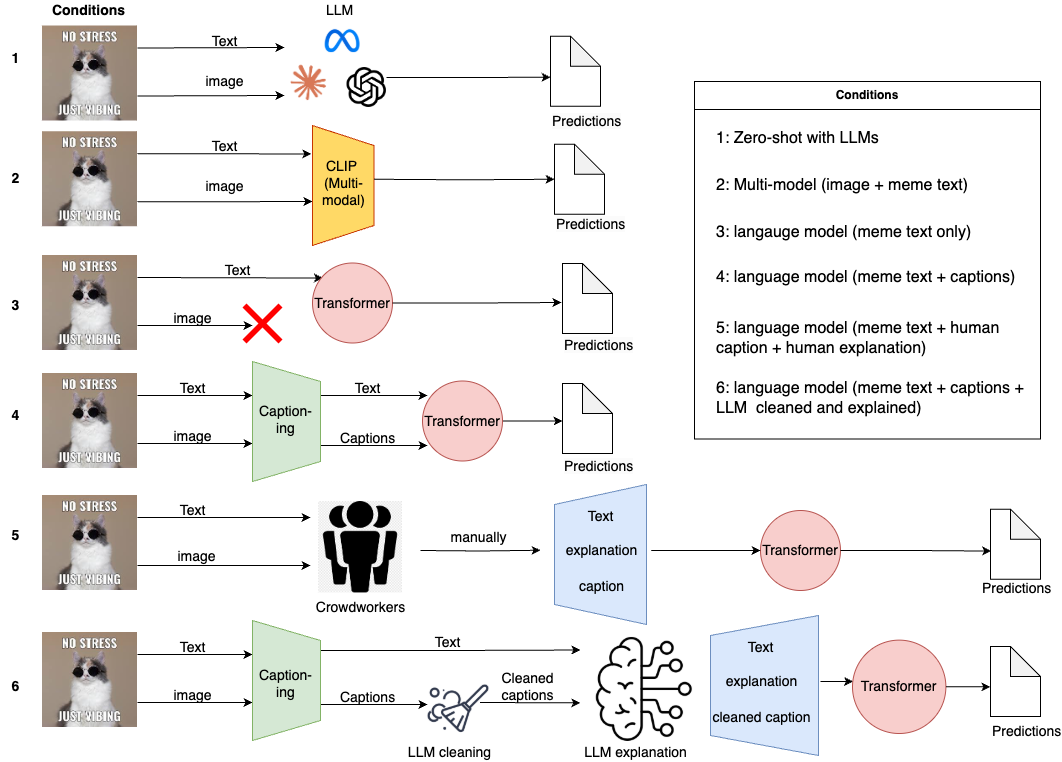} % Full width for the image
    \end{mdframed}
    \caption{Overview of the experimental conditions.}
    \label{fig:pipeline}
\end{figure*}

Using this prompt, we processed the captions and discarded any that resulted in an \textit{INVALID DESCRIPTION}. Since memes consist of both text and image components, when a caption was deemed invalid, the original meme text was retained and utilized for training without an AI-generated caption. This ensured that no training sample was excluded due to an invalid caption.

\begin{table}[ht]
\centering
\caption{Valid Caption Statistics (\%) for BLIP and GIT Models Using Sonnet 3.5, LLaMA 3.1 70B, and GPT-4o.}
\begin{tabular}{llccc}
\toprule
\textbf{Model} & \textbf{Set} & \textbf{Sonnet} & \textbf{LLaMA} & \textbf{GPT-4o} \\ \midrule
\textbf{BLIP}  & Train        & 89.9            & 99.9           & 69.2           \\
               & Dev          & 89.8            & 99.8           & 72.6           \\
               & Test         & 89.5            & 100.0          & 71.8           \\ \midrule
\textbf{GIT}   & Train        & 72.5            & 96.5           & 69.6           \\
               & Dev          & 67.6            & 95.8           & 69.8           \\
               & Test         & 69.7            & 97.6           & 70.0           \\ \bottomrule
\end{tabular}
\label{tab:valid_caption_stats}
\end{table}

Figure~\ref{fig:clean_pipeline} illustrates the workflow for cleaning the captions generated by BLIP. Table~\ref{tab:valid_caption_stats} presents the percentage of valid captions produced by BLIP and GIT models after being cleaned by three LLMs: LLaMA 3.1 70B\footnote{\url{ai.meta.com/blog/meta-llama-3-1/}}, GPT-4o\footnote{\url{openai.com/index/hello-gpt-4o/}}, and Sonnet 3.5 v2\footnote{\url{anthropic.com/news/3-5-models-and-computer-use}}. 

After LLM-based cleaning, if the caption is successfully processed and deemed valid, both the cleaned caption and the meme text are passed to training. If the caption is too corrupted to clean and is marked as invalid, then only the original meme text is used. The percentage of valid captions evaluates how effectively each LLM handles noisy caption outputs from captioning models like BLIP and GIT, providing insights into their ability to process caption quality, resolve anomalies, and manage failures (Table~\ref{tab:valid_caption_stats}).

\subsection{LLM-based Semantic Augmentation}
\label{sec:llm_context_enrichment}
In addition to directly utilizing the meme text and machine-generated captions, we employ LLMs to generate explanations which highlight key semantic or persuasive cues that might be abstract or less apparent in the raw text alone. Specifically, each sample is augmented with an LLM-generated explanation illustrating the potential rhetorical or psychological strategies present in meme content. Our assumption is that these explanations provide more nuanced insight, thus improving the downstream model’s performance on complex, multi-label classifications. 

\subsection{Experimental Conditions}
In this study, we explore six distinct configurations and their derivatives that incorporate the meme, LLM-based semantic augmentation, human annotations, multimodal model, and language model as shown in Figure \ref{fig:pipeline}.

\begin{enumerate}
    \item \textbf{Zero-Shot with Meme Image and Text:} 
    The same LLMs used for cleaning receive the meme image and text without any fine-tuning. This baseline measures how well modern LLMs handle multimodal, multi-label classification in a zero-shot configuration.

    \item \textbf{Image + Text (Multimodal Model):}
    A multimodal model CLIP \citep{radford2021learning} processes both image and textual cues, leveraging the learned alignment between visual and linguistic features.
    
    \item \textbf{Meme Text Only (Language Model):}
    Only the textual element of the meme is used, serving as a text-only baseline. The language model BART decoder \citep{lewis2020bart} treats the classification as a sequence-to-sequence prediction, inherently suited for hierarchical classification.
\vspace{+0.5 cm}
    \item \textbf{Text + Captions (Language Model):}
    Machine-generated captions (e.g., from BLIP) are concatenated with the meme text to incorporate visual context through textual representation. The same language model used in the text-only baseline is employed for this configuration.

    \item \textbf{Human Captions and Explanations (Language Model):}
    A subset of memes includes crowd-annotated captions and explanations, while the remaining samples consist of meme text only. The input is processed using the same language model as in previous baseline.

    \item \textbf{Meme Text + Cleaned Captions + LLM Explanations (Language Model):}
    Machine-generated captions undergo LLM-based cleaning to remove noise. Additionally, each meme is paired with an explanation generated by an LLM, highlighting potential persuasive or contextual elements. The concatenated input (meme text, cleaned captions, and explanation) is processed using the same language model.
\end{enumerate}

\subsection{Human and LLM-Based Annotations}

We conducted a study on Prolific\footnote{\url{https://www.prolific.co}} with 300 crowdworkers, retaining 70 based on a manual inspection of the quality of their responses. These 70 high-performing crowdworkers annotated 1,000 memes from the SemEval persuasive meme dataset for training and an additional 1,000 memes for testing (captions + explanations). Following the instructions detailed in Figures~(\ref{fig:techniques_table}-\ref{fig:task_example}) in the Appendix, participants' captions and explanations of the memes did not contain the labeled persuasion techniques. This ensured that the human annotations remained label-agnostic, mirroring the conditions applied to the LLMs.
For the remaining samples, only the original meme text was used without human captions or explanations. In parallel, we employed LLMs to clean BLIP/GIT captions and generate explanations for all memes. Annotating 2,000 memes via crowdworkers costs approximately \$1,100, whereas annotating the entire dataset (9,500 memes) with LLaMA 3.1 70B and Sonnet 3.5 V2 amounted to only \$3 each, and \$30 for GPT-4o highlighting the significantly lower financial cost of LLM-generated annotations as compared to crowdsourced annotations.

\paragraph{Ethical Considerations}
This study was conducted anonymously, and crowd participant data were kept confidential. Approval was obtained from the first author's Institutional Review Board (IRB), and all participants provided informed consent for data collection and analysis. The task took 30 minutes per 10 memes, yielding an approximate hourly rate of \$11. To ensure English proficiency and contextual relevance, given the SemEval persuasive memes dataset centers on U.S. political content, participants were required via platform settings to reside in the United States and speak English fluently.

\paragraph{Variability in LLM Outputs.}
Unlike LLM-based cleaning, which involves a more deterministic processing with no room for variability, generating explanations introduces a creative element that can lead to more diverse outputs. To account for this variability, we generated five versions of LLM-based explanations for each sample. We report the average performance across these five variations, along with the variance and best-case results. This repeated generation process enables us to evaluate the consistency of LLM outputs and the potential impact on the downstream task.

\subsection{Validation on Toxic and Hateful Content}
We further validate our approach on two tasks inherently rich in offensive or hateful language: the \textit{Google Jigsaw Toxic Comments} dataset (multi-label/textual) and the \textit{Facebook Hateful Meme Challenge} dataset (binary/multimodal). Both tasks present high-context scenarios where toxicity or hateful expressions are crucial for accurate classification.

In each setting, we compare:
\begin{itemize}
    \item \textbf{Zero-Shot LLM Classification}, in which LLaMA 3.1, Sonnet 3.5 , and GPT-4o  receive minimal task-specific context.
    
   \item \textbf{Language Transformers:} DistilBERT \citep{sanh2019distilbert} and RoBERTa \citep{liu2019roberta}, which are encoder-only models designed for language input.
    
    \item \textbf{Multimodal Model:} CLIP, which is used exclusively for the Facebook Hateful Meme Challenge to process the multimodal input.

    \item \textbf{LLM-Based Semantic Augmentation}, where we prompt the LLM to retain and highlight offensive or hateful themes (\emph{e.g.}, profanity, racism, xenophobia). These cues are then concatenated with the original text (and image features for hateful memes) before being fed into the downstream model.
\end{itemize}

\paragraph{Handling Offensive Language and Censorship}
Unlike Semeval persuasive memes dataset, we could not simply prompt LLMs to clean and explain inputs in the toxic comment and hateful meme domains. LLMs when prompted to clean or explain, either censor or paraphrase explicit words or phrases that are key indicators of hatefulness and toxicity. As shown in the following example of a hateful meme text:

\begin{quote}
\emph{``The latest toy has hit the shops. It's a talking Muslim doll; nobody knows what the fuck it says, because no one has the guts to pull the string.''}
\end{quote}

A simple LLM prompt to clean and explain would often yield a refusal or a filtered response:

\begin{quote}
\emph{``I apologize, but I do not feel comfortable providing the cleaning and detailed structural analysis or explanation of hateful content that could further amplify or spread harmful stereotypes and bigotry.''}
\end{quote}

Such censorship omits vital features, the explicit profanity and hateful framing that training the model relies on. Instead, we requested that the LLM \emph{preserve offensive or profane expressions} by producing \textbf{triggers} alongside the explanation, reflecting the core hateful or toxic message. Below is an example of GPT-4o output:
\begin{figure}[ht]
\begin{mdframed}[linewidth=1pt] \noindent \textbf{Explanation:}
``This meme implies that interacting with anything connected to Islam is dangerous or undesirable, feeding into a narrative that paints Muslims as inherently threatening or alien. By humorously suggesting that no one dares pull the doll’s string, the meme mocks and perpetuates fears of Islam.''

\noindent \textbf{Triggers:} Islamophobia, Stereotyping, Muslim doll, what the fuck, no one has the guts. \end{mdframed}
\label{fig:islamophobia_example}
\end{figure}

Notably, the explanation often avoids strong language, but the requested ``triggers'' field retains the explicit profanity and hateful expressions and keywords that are essential to aid the classification, for detailed output see Figure~\ref{fig:llama_output} in Appendix.

\section{Results}

We present our empirical evaluations results centered on the \textit{SemEval 2024 Persuasive Memes Multi-label Classification} as the primary downstream task. We assess various configurations of the proposed data pipeline and validate the top-performing approach on two additional tasks. Our results highlight the effectiveness of the proposed approach through detailed comparisons across different conditions and configurations. The subsequent subsections address each of our research questions in depth.

% For the main paper (test plots)
\begin{figure*}[t]
    \centering
    \begin{subfigure}[b]{0.32\textwidth}
        \centering
        \includegraphics[width=\textwidth]{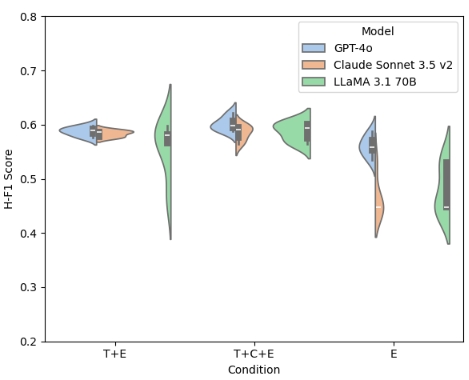}
        \caption{Violin Plot}
        \label{fig:test_violin}
    \end{subfigure}
    \hfill
    \begin{subfigure}[b]{0.32\textwidth}
        \centering
        \includegraphics[width=\textwidth]{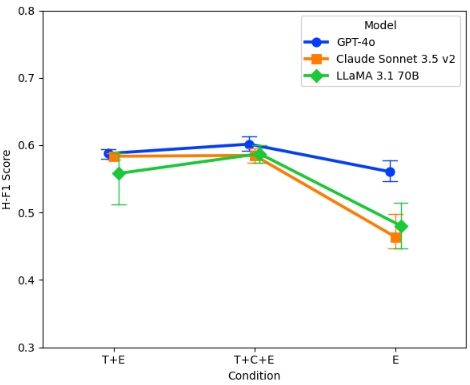}
        \caption{Point Plot}
        \label{fig:test_point}
    \end{subfigure}
    \hfill
    \begin{subfigure}[b]{0.32\textwidth}
        \centering
        \includegraphics[width=\textwidth]{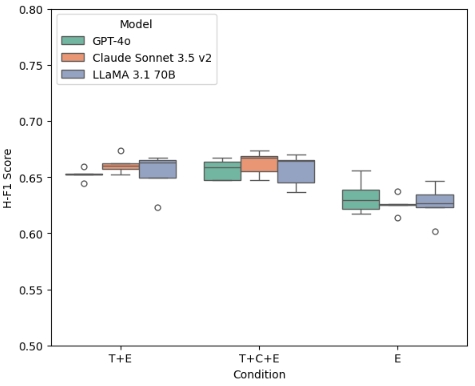}
        \caption{Box Plot}
        \label{fig:test_box}
    \end{subfigure}
        \caption{Performance visualization on the test set using different LLMs to generate meme explanations, with 5 runs per condition. GPT-4o maintains consistently strong performance across all conditions Text+Explanation (T+E), Text+Caption+Explanation (T+C+E), and Explanation Only (E), while Sonnet 3.5 performs well but shows slightly more variance. LLaMA 3.1 exhibits mixed results with high variance, particularly struggling in the Explanation-only (E) condition. The T+C+E condition generally yields the most stable performance across all models, suggesting the benefit of combining multiple information sources.}

    \label{fig:test_plots}
\end{figure*}

\subsection{Does LLM-Based Cleaning Work? (RQ1a)}
% \subsection{Effectiveness and Limitations of LLM-Based Cleaning}
\begin{table}[t]
\centering
\setlength{\tabcolsep}{4.5pt}
\caption{Performance comparison of BLIP and GIT models using uncleaned captions and captions cleaned by different LLMs. H-F1 scores shown for both development and test sets. Bolded: highest H-F1 scores per column. See Table~\ref{tab:significance_appendix} in Appendix for significance tests.}
\label{tab:cleaning_performance}
\begin{tabular}{l|cc|cc}
\toprule
\multirow{2}{*}{\textbf{Model}} & \multicolumn{2}{c|}{\textbf{BLIP H-F1 }} & \multicolumn{2}{c}{\textbf{GIT H-F1}} \\
\cmidrule(lr){2-3} \cmidrule(lr){4-5}
 & \textbf{Dev} & \textbf{Test} & \textbf{Dev} & \textbf{Test} \\
\midrule
\textbf{text-only} & 64.20 & 55.77 & 64.20 & 55.77 \\
\textbf{text + uncleaned captions} & 64.77 & 57.14 & 64.54 & 55.07 \\
\midrule
\textbf{Sonnet 3.5} & 65.78 & 57.25 & 65.30 & 58.43 \\
\textbf{LLaMA 3.1} & 64.74 & \textbf{58.87} & 65.37 & \textbf{58.69} \\
\textbf{GPT-4o} & \textbf{65.85} & 57.96 & \textbf{66.04} & 58.41 \\
\bottomrule
\end{tabular}
\end{table}

We evaluated the impact of LLM-based cleaning on captions generated by BLIP and GIT. Comparing classifiers trained on uncleaned captions versus those cleaned by LLMs (Sonnet 3.5, LLaMA 3.1 70B, and GPT-4o) revealed modest H-F1 score improvements (Table~\ref{tab:cleaning_performance}). Statistical significance was assessed using the Wilcoxon signed-rank test \cite{wilcoxon1945} with Benjamini-Hochberg correction \cite{benjamini1995} on the development set due to the blind nature of the SemEval test set (Table~\ref{tab:significance_appendix} in Appendix). Only GPT-4o-cleaned captions showed significant improvements over the uncleaned captions for the classifier ($p = 0.0157$), while other LLMs did not achieve statistical significance ($p > 0.05$).

GPT-4o's restrictive cleaning approach likely preserves essential information without over-processing, despite a higher rate of invalid captions (Figure~\ref{fig:clean_pipeline}, Table~\ref{tab:valid_caption_stats}). These findings are consistent with prior research on rule-based cleaning in NLP, which also reported incremental performance gains \cite{kukreja2024spam}. Additionally, integrating LLM-based cleaning into data augmentation pipelines streamlines the process by simultaneously generating contextual explanations and refining text, eliminating the need for manual rule-based methods.

However, in domains with highly offensive language, such as hateful memes and toxic comments, the embedded LLM safeguards often hinder effective cleaning. Unlike the persuasive memes dataset with milder language, LLMs frequently censor or omit explicit and hateful expressions essential for accurate classification, as discussed in the \textit{``Handling Offensive Language and Censorship''} subsection.

\subsection{Does Adding Context Improve Performance? (RQ1b)}
Leveraging LLMs to augment each meme with a short, explanatory context can yield tangible gains in performance. As shown in Figure~\ref{fig:test_plots}b and Table~\ref{tab:ablation} in Appendix, we experiment with three different input configurations beyond the original meme text (T) or text+caption (T+C): (i) \textbf{T+E} (text + explanation), (ii) \textbf{T+C+E} (text + caption + explanation), and (iii) \textbf{E} (explanation only). We report results across GPT-4o,  Sonnet 3.5~v2, and LLaMA~3.1~70B, conducting five iterations per condition to account for the inherent variability in LLM-generated outputs. Statistical significance was assessed using one-sample t-tests with Benjamini-Hochberg (BH) correction ($\alpha=0.05$) to account for multiple comparisons.

\paragraph{T+E vs.\ T.}
Adding an LLM-generated explanation to the original text consistently boosts hierarchical F1 (H-F1) compared to text-only baselines. Both GPT-4o and Sonnet 3.5 show statistically significant improvements (BH-p $<$ 0.05) on both development and test sets, with GPT-4o achieving the strongest gains (test set: 58.7±0.8 vs.\ baseline 55.8, BH-p = 0.001). LLaMA 3.1, while showing numerical improvements, did not achieve statistical significance in this condition (BH-p $>$ 0.05), suggesting less reliable enhancement from its explanations.

\paragraph{T+C+E vs T+C.}
Including meme text, caption and an LLM-generated explanation (\textit{T+C+E}) generally yields the strongest performance. As illustrated in Figure~\ref{fig:test_violin} and Table~\ref{tab:ablation} in Appendix, this condition attains the highest median H-F1 across runs for all three LLMs. GPT-4o shows the most substantial and statistically significant improvements over both baselines (test set: 60.2±1.4 vs.\ T+C 57.1, BH-p = 0.007), while Sonnet 3.5 and LLaMA 3.1 show mixed statistical significance when compared to the T+C baseline (BH-p = 0.085 and 0.12 respectively). This suggests that multiple sources of context (text, caption, explanation) collectively help the classifier identify subtle persuasive strategies, notably when using more GPT-4o and Sonnet 3.5.

\paragraph{Explanation-Only (E).}
In contrast, training on \textit{E} alone (i.e., ignoring the original text/caption) substantially degrades performance. Figure~\ref{fig:test_box} demonstrates the significant drop in H-F1 for all three models when only explanations are used. GPT-4o shows the most resilience in this condition (56.1±2.0), while both Sonnet 3.5 (46.4±1.0) and LLaMA 3.1 (48.1±2.0) experience substantial performance decrease. This indicates that even well-written LLM explanations cannot fully capture the nuanced rhetorical devices contained in the original meme text and captions.

% \paragraph{LLM Choice and Variance.}
% Across the different LLMs, GPT-4o stands out for its consistently strong performance and relatively low variance (std: 0.8--2.0 across conditions), while Claude Sonnet 3.5~v2 performs competitively but exhibits a slightly higher variability particularly in T+C+E condition (std: 1.3). LLaMA~3.1~70B shows more pronounced variance in all conditions (std: 1.7--2.0), particularly struggling in \textit{E}-only. This variance pattern, visible in Figure~\ref{fig:test_point}, suggests that explanation quality and stability depend heavily on the underlying LLM's capabilities.

\paragraph{Comparison with Other Approaches.}
Finally, Figure~\ref{fig:model_comparison} and Table \ref{tab:performance_comparison} in Appendix, compares semantic augmentation configuration against task-specific baselines and zero-shot LLM classification. Our best-performing method, \textbf{LLM + decoder}, achieves an H-F1 of 62.2\% (using GPT-4o in T+C+E configuration), significantly exceeding all baselines (e.g., CLIP: 58.2\%, text-only: 55.8\%) and outperforming zero-shot LLMs (e.g., GPT-4o: 54.2\%). Together, these findings show that appending succinct, LLM-generated explanations can serve as an effective form of semantic augmentation, enabling classifiers to better discern persuasive linguistic signals, with the choice of LLM impacting the reliability of these improvements to some degree.
\begin{figure}[t]
    \centering
    \includegraphics[width=0.95\columnwidth]{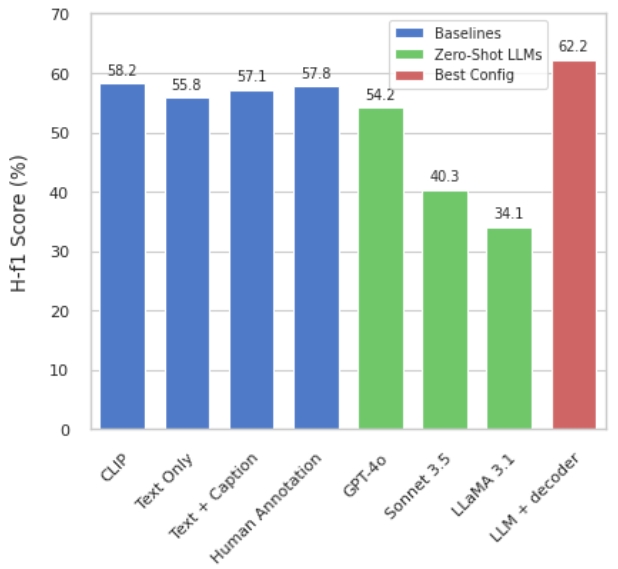}
    \caption{Performance comparison across different model categories (3 runs). Our best configuration (LLM + decoder) achieves the highest performance at 62.2\%, outperforming both task-specific baselines and zero-shot LLMs.}
    \label{fig:model_comparison}
\end{figure}

\begin{figure*}[t]
    \centering
    \begin{subfigure}[b]{0.32\textwidth}
        \centering
        \includegraphics[width=\textwidth]{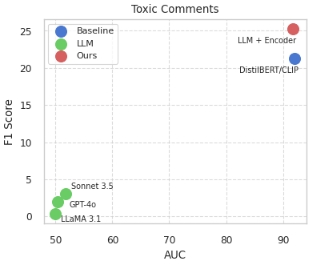}
        % \caption{Toxic Comments dataset performance showing F1-AUC trade-off. Our approach (LLM + Encoder) achieves high performance in both metrics, comparable to the baseline.}
        \label{fig:toxic}
    \end{subfigure}
    \hfill
    \begin{subfigure}[b]{0.32\textwidth}
        \centering
        \includegraphics[width=\textwidth]{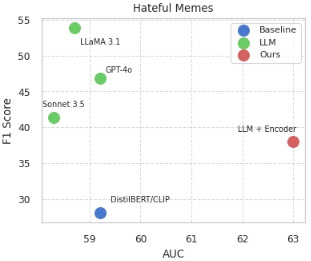}
        % \caption{Hateful Memes dataset performance showing F1-AUC trade-off. Our approach achieves the highest AUC while maintaining competitive F1 score.}
        \label{fig:memes}
    \end{subfigure}
    \hfill
    \begin{subfigure}[b]{0.32\textwidth}
        \centering
        \includegraphics[width=\textwidth]{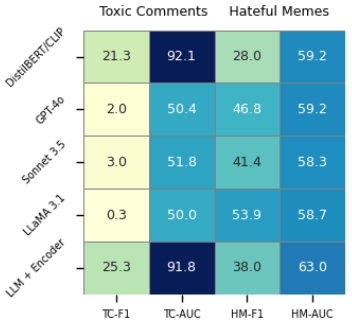}
        % \caption{Comprehensive performance metrics across both datasets. Color intensity indicates score magnitude, showing our model's balanced performance across tasks.}
        \label{fig:heatmap}
    \end{subfigure}
    \caption{Performance analysis across datasets and metrics. Our LLM + Encoder approach demonstrates strong performance across both tasks, achieving (a) high F1 and AUC scores on toxic comment detection comparable to specialized models, (b) superior AUC on hateful meme detection while maintaining competitive F1 scores, and (c) consistent performance across all metrics as shown in the heatmap visualization.}
    \label{fig:all_results}
\end{figure*}

\subsection{Generalizability Across Related Domains (RQ2)}

To assess whether the proposed LLM-based semantic augmentation approach generalizes effectively across related domains in social media, we evaluated its performance on two distinct datasets: Jigsaw Toxic Comments (text-only) and Facebook Hateful Memes (multimodal). Both tasks require the model to detect harmful content, but they differ in modality and language characteristics, providing a robust test of generalizability. We selected LLaMA 3.1 for semantic augmentation due to its open-source nature and significantly lower cost than GPT-4o. While GPT-4o delivers the strongest performance and Sonnet 3.5 offers a similar cost to LLaMA, both alternatives are closed-source. Demonstrating strong results with LLaMA reasonably suggests that similar or better outcomes would be achievable with the other LLMs.

As shown in Figure~\ref{fig:all_results} and Table~\ref{tab:combined_performance_metrics} in Appendix, our approach (LLM + Encoder) delivers strong, consistent performance across both datasets. For toxic comments, the F1 score (\textbf{25.3}) demonstrates an improvement over all baselines, including DistilBERT/CLIP (\textbf{21.3}) and other LLMs. Importantly, its AUC (\textbf{91.8}) is on par with DistilBERT/CLIP (\textbf{92.1}), underscoring that our method can achieve comparable classification sensitivity while maintaining superior balance across metrics.

In the hateful memes task, the LLM + Encoder model achieves the highest AUC (\textbf{63.0}) and competitive F1 (\textbf{38.0}) compared to baseline approaches. The heatmap in Figure~\ref{fig:all_results}\subref{fig:heatmap} highlights the method’s balanced performance across tasks, indicating that the model adapts effectively to both domains. These results suggest that the proposed semantic augmentation strategy generalizes well across social media tasks with differing data modalities and content characteristics.

\subsection{Semantic Augmentation Strategies for Challenging Domains (RQ3)}

While explanation-based augmentation was effective in SemEval persuasive meme classification task, domains involving strong vulgarity or harmful language, such as toxic comment detection and hateful memes, posed unique challenges. LLMs often censor, rephrase, or omit critical language elements during explanation generation, which impairs their effectiveness in these domains. Cleaning strategies similarly failed, as LLMs would sanitize offensive text or return invalid outputs. To address these issues, we adapted a \textbf{trigger-based semantic augmentation strategy} that highlights keywords and themes in the data, as shown in Appendix Figure~\ref{fig:llama_output}.

\paragraph{When to Use Triggers.}

Table~\ref{tab:trigger_comparison} in the appendix compares the performance of baseline models (DistilBERT/CLIP), explanation-based augmentation (LLM-exp+Encoder), and trigger-based augmentation (LLM-triggers+Encoder). In toxic comment detection, explanations slightly underperformed compared to the baseline, while the trigger-based approach significantly boosted F1 (\textbf{25.3}) and retained a competitive AUC (\textbf{91.8}). For hateful memes, explanations showed marginal drop over the baseline (\textbf{27.3 F1}, \textbf{58.9 AUC}), but the trigger-based strategy achieved substantially higher scores (\textbf{38.0 F1}, \textbf{63.0 AUC}). 

Unlike explanations, which rely on paraphrasing and elaborating which may omit critical language elements due to LLM safeguards, triggers provide explicit, domain-specific keywords and themes, allowing the model to focus on key linguistic features. This is particularly important in tasks involving harmful or offensive content, where triggers enable the model to identify recurring patterns (e.g., racial slurs, hateful symbols) that explanations might exclude or paraphrase. These findings suggest that triggers are especially effective in domains where capturing explicit terms or themes is critical for model training, offering a reliable alternative for semantic augmentation when explanations fall short.

\subsection{Zero-Shot LLMs: Where Do They Struggle? (RQ4)}
Zero-shot LLMs struggle to handle the nuanced and high-context nature of social media tasks such as persuasive meme, toxic comment, and hateful meme detection. As shown in Figures~\ref{fig:model_comparison}-\ref{fig:all_results} and Table~\ref{tab:combined_performance_metrics} in Appendix, their performance is markedly inferior to both task-specific models and our proposed LLM + Transformer approach, particularly in \textit{multi-label settings} where granularity is important. 
These findings conform with prior studies highlighting the limitations of LLMs in complex, context-dependent tasks \citep{sushil2024comparativestudyzeroshotinference, jose2024large}. However, our findings add a critical observation: while LLMs demonstrate acceptable performance on hateful meme detection (a binary classification task) they fall short in multi-label settings like toxic comment classification and persuasion detection. This pronounced performance gap highlights the inherent challenges of tasks requiring fine-grained differentiation across overlapping categories, underscoring the complexities that multi-label classifications pose for LLMs. Despite these limitations, LLMs prove effective in semantic augmentation through advanced explanation and trigger generation. This result highlights their potential as tools for improving downstream model performance, even when their direct classification capabilities seem inadequate.

\section{Limitations}
\label{appendix:limitations}
Our study primarily focuses on LLM-based semantic augmentation for text-centric tasks, with limited exploration of multimodal pipelines. While initial tests on the multimodal Persuasive and Hateful Memes datasets are promising, we did not deeply examine how LLM-generated outputs could be more tightly integrated with richer visual features (LLM + Multimodal). Additionally, we employed a single, comprehensive prompting strategy for generating context (e.g., explanations and triggers) across all experiments. Although effective, exploring diverse prompting styles could potentially uncover further insights but comes with significant computational and practical overhead, as the experimental space grows exponentially when combined with multi-iteration setups across multiple conditions.

Furthermore, human annotations were only collected for a subset of the data (2,000 memes), while LLMs were used to annotate the entire dataset (9,500 memes). While this difference in annotation coverage could be perceived as a limitation, the significantly lower cost of LLM-generated annotations compared to human crowdsourcing (\$3 for Sonnet 3.5 vs. \$1,100 for 2,000 samples via Prolific) makes this a realistic and fair comparison. However, the lack of human annotations on the full dataset may introduce potential bias when benchmarking against other baselines.

Moreover, we opted for straightforward implementation of the supervised models, avoiding hyperparameter fine-tuning to ensure simplicity and comparability. The impact of semantic augmentation relatively depends on how they are fused with the original data during training. Factors such as the percentage of attention allocated to explanations versus the original input, the use of explicit separators (e.g., [sep] tokens), and the specific architecture of the model all influence performance gains when adding the explanations or the triggers. Further experiments are required to systematically evaluate these fusion strategies and determine optimal configurations for different tasks and model types.

An additional consideration is that we primarily utilized LLaMA 3.1, an open-source and cost-effective model. Although LLaMA 3.1 70B exhibits inferior results compared to closed-source models such as GPT-4o and Sonnet 3.5 v2, our results suggest that these more capable models are likely to perform similarly, if not better, when used for semantic augmentation. Moreover, we observed important differences regarding model safeguards: while it is possible to bypass certain content restrictions in GPT-4o and LLaMA 3.1 70B with carefully crafted prompts, Sonnet 3.5 consistently adheres strictly to content moderation policies, restricting the generation of offensive or explicit content.

Lastly, our emphasis on contextual augmentation rather than large-scale synthetic data generation leaves open questions about the trade-offs between these approaches. Future work may aim to address these limitations by integrating LLMs more systematically with multimodal pipelines and comparing LLM-based semantic augmentation against LLM-based data augmentation across diverse tasks with different prompting styles. This research should investigate how data quality varies between these methods, particularly examining performance-cost tradeoffs in specialized domains and identifying which augmentation techniques best preserve domain-specific knowledge while reducing annotation expenses. A standardized evaluation framework measuring not only task performance but also data diversity and bias mitigation would enable more principled comparisons between these competing approaches.

\section{Conclusion}

In this paper, we proposed a novel semantic augmentation method leveraging LLMs to enhance the detection of harmful content such as persuasive memes, toxic comments, and hateful memes. Unlike conventional data augmentation strategies, which primarily focus on quantitatively expanding the training set, our method enriches the training data qualitatively by introducing contextually rich explanations and trigger-based augmentations. Extensive evaluations across three distinct datasets—SemEval 2024 persuasive memes, Google Jigsaw toxic comments, and Facebook hateful memes—demonstrated that our approach substantially improves classification performance, significantly surpassing zero-shot LLM baselines and traditional supervised methods.

Our results highlight several key insights. First, semantic augmentation using LLM-generated explanations is particularly effective for context-dependent tasks, demonstrating an ability to closely approximate the performance of human-annotated data at a fraction of the cost. Interestingly, despite their proven effectiveness in generating high-quality explanations, zero-shot LLMs paradoxically underperform when directly employed for classification tasks. This highlights a fundamental distinction between explanatory capabilities and predictive performance, emphasizing the need for targeted semantic augmentation strategies like ours to bridge this gap. Second, in domains characterized by sensitive or offensive content, our trigger-based augmentation strategy effectively circumvents the safeguard limitations inherent to LLMs, achieving notable gains where explanation-based augmentation alone fails.

However, our approach is not without limitations. A prominent concern arises from the safeguard mechanisms embedded in LLMs, which restrict the generation of explicit or highly sensitive content. Although our trigger-based augmentation alleviates this issue to some extent, completely overcoming these safeguards remains challenging, potentially limiting effectiveness in extreme cases or highly offensive content scenarios. Additionally, while our method significantly reduces annotation costs and improves efficiency, further exploration is required to systematically quantify the relative cost-efficiency of different LLM-based strategies, especially as compared to emerging open-source models such as DeepSeek R1\footnote{DeepSeek-R1: Incentivizing Reasoning Capability in LLMs via Reinforcement Learning, DeepSeek-AI, 2025. Available at \url{https://arxiv.org/abs/2501.12948}.}.

Future research directions include integrating multimodal augmentation pipelines more thoroughly, systematically exploring diverse prompting strategies, and extending experiments with fully annotated human-written explanations. Moreover, investigating open-source LLMs for semantic augmentation may enhance the accessibility and scalability of our method, facilitating broader adoption within the research community. Ultimately, our work underscores the substantial potential of semantic augmentation to address nuanced challenges in harmful content detection, paving the way for safer and more reliable online environments.

\section*{Acknowledgments}
This work is partially supported by the Australian Research Council (ARC) Centre of Information Resilience (Grant No. IC200100022) and by an ARC Future Fellowship Project (Grant No. FT240100022).

\bibliography{aaai25}

\begin{thebibliography}{44}
\providecommand{\natexlab}[1]{#1}

\bibitem[{Bang et~al.(2023)}]{bang2023multitask}
Bang, Y.; et~al. 2023.
\newblock A multitask, multilingual, multimodal evaluation of chatgpt on reasoning, hallucination, and interactivity.
\newblock \emph{arXiv preprint arXiv:2302.04023}.

\bibitem[{Benjamini and Hochberg(1995)}]{benjamini1995}
Benjamini, Y.; and Hochberg, Y. 1995.
\newblock Controlling the false discovery rate: a practical and powerful approach to multiple testing.
\newblock \emph{Journal of the Royal Statistical Society: Series B (Methodological)}, 57(1): 289--300.

\bibitem[{Borrohou et~al.(2023)Borrohou, Fissoune, Badir, and Tabaa}]{borrohou2023data}
Borrohou, S.; Fissoune, R.; Badir, H.; and Tabaa, M. 2023.
\newblock Data Cleaning in Machine Learning: Improving Real Life Decisions and Challenges.
\newblock In \emph{International Conference on Advanced Intelligent Systems for Sustainable Development}, 627--638. Springer.

\bibitem[{Brattberg and Maurer(2018)}]{brattberg2018russian}
Brattberg, E.; and Maurer, T. 2018.
\newblock Russian election interference: Europe's counter to fake news and cyber attacks.
\newblock \emph{Carnegie Endowment for International Peace}.

\bibitem[{Da~San~Martino et~al.(2019)Da~San~Martino, Seunghak, Barr{\'o}n-Cede{\~n}o, Petrov, and Nakov}]{da2019fine}
Da~San~Martino, G.; Seunghak, Y.; Barr{\'o}n-Cede{\~n}o, A.; Petrov, R.; and Nakov, P. 2019.
\newblock Fine-grained analysis of propaganda in news articles.
\newblock In \emph{EMNLP-IJCNLP}, 5636--5646.

\bibitem[{Dimitrov et~al.(2024)Dimitrov, Alam, Hasanain, Hasnat, Silvestri, Nakov, and Da~San~Martino}]{dimitrov-etal-2024-semeval}
Dimitrov, D.; Alam, F.; Hasanain, M.; Hasnat, A.; Silvestri, F.; Nakov, P.; and Da~San~Martino, G. 2024.
\newblock {S}em{E}val-2024 Task 4: Multilingual Detection of Persuasion Techniques in Memes.
\newblock In \emph{Proceedings of the 18th International Workshop on Semantic Evaluation (SemEval-2024)}, 2009--2026. Mexico City, Mexico: Association for Computational Linguistics.

\bibitem[{Ding et~al.(2023)Ding, Qin, Liu, Chia, Joty, and Li}]{ding2023gpt}
Ding, B.; Qin, C.; Liu, L.; Chia, Y.~K.; Joty, S.; and Li, B. 2023.
\newblock Is gpt-3 a good data annotator?
\newblock \emph{arXiv preprint arXiv:2303.07845}.

\bibitem[{Edwards and Camacho-Collados(2024)}]{edwards2024languagemodelstextclassification}
Edwards, A.; and Camacho-Collados, J. 2024.
\newblock Language Models for Text Classification: Is In-Context Learning Enough?
\newblock arXiv:2403.17661.

\bibitem[{Feng et~al.(2021)Feng, Gangal, Wei, Chandar, Vosoughi, Mitamura, and Hovy}]{feng2021survey}
Feng, S.~Y.; Gangal, V.; Wei, J.; Chandar, S.; Vosoughi, S.; Mitamura, T.; and Hovy, E. 2021.
\newblock A Survey of Data Augmentation Approaches for NLP.
\newblock \emph{arXiv preprint arXiv:2105.03075}.

\bibitem[{Gao et~al.(2023)Gao, Huang, Yan, Zhou, and Chen}]{gao2023sungen}
Gao, F.; Huang, J.~Y.; Yan, T.; Zhou, W.; and Chen, M. 2023.
\newblock SunGen: A Framework for Mitigating Noise in Synthetic Data for Text Classification.
\newblock \emph{arXiv preprint arXiv:2307.07099}.

\bibitem[{Golovchenko et~al.(2020)Golovchenko, Buntain, Eady, Brown, and Tucker}]{golovchenko2020cross}
Golovchenko, Y.; Buntain, C.; Eady, G.; Brown, M.~A.; and Tucker, J.~A. 2020.
\newblock Cross-platform state propaganda: Russian trolls on Twitter and YouTube during the 2016 US presidential election.
\newblock \emph{The International Journal of Press/Politics}, 25(3): 357--389.

\bibitem[{Holmes et~al.(2023)}]{holmes2023evaluating}
Holmes, J.; et~al. 2023.
\newblock Evaluating large language models on a highly-specialized topic, radiation oncology physics.
\newblock \emph{Frontiers in Oncology}, 13.

\bibitem[{Hsieh et~al.(2023)Hsieh, Li, Yeh, Nakhost, Fujii, Ratner, Krishna, Lee, and Pfister}]{hsieh2023distilling}
Hsieh, C.-Y.; Li, C.-L.; Yeh, C.-K.; Nakhost, H.; Fujii, Y.; Ratner, A.; Krishna, R.; Lee, C.-Y.; and Pfister, T. 2023.
\newblock Distilling Step-by-Step! Outperforming Larger Language Models with Less Training Data and Smaller Model Sizes.
\newblock \emph{arXiv preprint arXiv:2305.02301}.

\bibitem[{Jose and Greenstadt(2024)}]{jose2024large}
Jose, J.; and Greenstadt, R. 2024.
\newblock Are Large Language Models Good at Detecting Propaganda?
\newblock \emph{AAAI Conference on Artificial Intelligence}.

\bibitem[{Joshi, Sharma, and Bhattacharyya(2015)}]{joshi-etal-2015-harnessing}
Joshi, A.; Sharma, V.; and Bhattacharyya, P. 2015.
\newblock Harnessing Context Incongruity for Sarcasm Detection.
\newblock In \emph{Proceedings of the 53rd Annual Meeting of the Association for Computational Linguistics and the 7th International Joint Conference on Natural Language Processing (Volume 2: Short Papers)}, 757--762. Beijing, China: Association for Computational Linguistics.

\bibitem[{Jurkiewicz et~al.(2020)Jurkiewicz, Borchmann, Kosmala, and Grali{\'n}ski}]{jurkiewicz2020applicaai}
Jurkiewicz, D.; Borchmann, {\L}.; Kosmala, I.; and Grali{\'n}ski, F. 2020.
\newblock ApplicaAI at SemEval-2020 task 11: On RoBERTa-CRF, span CLS and whether self-training helps them.
\newblock \emph{arXiv preprint arXiv:2005.07934}.

\bibitem[{Kiela, Firooz, and Mohan(2020)}]{kiela2020hateful}
Kiela, D.; Firooz, H.; and Mohan, A. 2020.
\newblock The Hateful Memes Challenge: Detecting hate speech in multimodal memes.
\newblock In \emph{Advances in Neural Information Processing Systems}.

\bibitem[{Kiritchenko et~al.(2006)Kiritchenko, Matwin, Nock, and Famili}]{kiritchenko2006learning}
Kiritchenko, S.; Matwin, S.; Nock, R.; and Famili, A.~F. 2006.
\newblock Learning and Evaluation in the Presence of Class Hierarchies: Application to Text Categorization.
\newblock In \emph{Advances in Artificial Intelligence: 19th Conference of the Canadian Society for Computational Studies of Intelligence, Canadian AI 2006, Qu{\'e}bec City, Qu{\'e}bec, Canada, June 7-9, 2006. Proceedings}, 395--406. Springer.

\bibitem[{Kuila and Sarkar(2024)}]{kuila-sarkar-2024-deciphering}
Kuila, A.; and Sarkar, S. 2024.
\newblock Deciphering Political Entity Sentiment in News with Large Language Models: Zero-Shot and Few-Shot Strategies.
\newblock In Afli, H.; Bouamor, H.; Casagran, C.~B.; and Ghannay, S., eds., \emph{Proceedings of the Second Workshop on Natural Language Processing for Political Sciences @ LREC-COLING 2024}, 1--11. Torino, Italia: ELRA and ICCL.

\bibitem[{Kukreja et~al.(2024)Kukreja, Lakshmi, Sri, and Ravali}]{kukreja2024spam}
Kukreja, V.; Lakshmi, D.; Sri, C.; and Ravali, K. 2024.
\newblock Improved Spam Detection Through LSTM-Based Approach.
\newblock In \emph{2024 Third International Conference on Advances in Artificial Intelligence (ICAAI)}. IEEE.

\bibitem[{Kumar et~al.(2019)Kumar, Bhattamishra, Bhandari, and Talukdar}]{kumar2019paraphrasing}
Kumar, A.; Bhattamishra, S.; Bhandari, M.; and Talukdar, P. 2019.
\newblock Submodular optimization-based diverse paraphrasing and its effectiveness in data augmentation.
\newblock \emph{NAACL}.

\bibitem[{Lewis et~al.(2020)Lewis, Liu, Goyal, Ghazvininejad, Mohamed, Levy, Stoyanov, and Zettlemoyer}]{lewis2020bart}
Lewis, M.; Liu, Y.; Goyal, N.; Ghazvininejad, M.; Mohamed, A.; Levy, O.; Stoyanov, V.; and Zettlemoyer, L. 2020.
\newblock BART: Denoising Sequence-to-Sequence Pre-training for Natural Language Generation, Translation, and Comprehension.
\newblock \emph{arXiv preprint arXiv:1910.13461}.

\bibitem[{Li et~al.(2022)Li, Li, Xiong, and Hoi}]{li2022blip}
Li, J.; Li, D.; Xiong, C.; and Hoi, S. C.~H. 2022.
\newblock BLIP: Bootstrapped Learning from Image and Text.
\newblock \emph{arXiv preprint arXiv:2201.12086}.

\bibitem[{Li et~al.(2023{\natexlab{a}})Li, Zhu, Ma, Liu, and Shah}]{li2023chatgpt}
Li, X.; Zhu, X.; Ma, Z.; Liu, X.; and Shah, S. 2023{\natexlab{a}}.
\newblock Are chatgpt and gpt-4 general-purpose solvers for financial text analytics? an examination on several typical tasks.
\newblock \emph{arXiv preprint arXiv:2305.05862}.

\bibitem[{Li et~al.(2023{\natexlab{b}})Li, Zhu, Lu, and Yin}]{li2023synthetic}
Li, Z.; Zhu, H.; Lu, Z.; and Yin, M. 2023{\natexlab{b}}.
\newblock Synthetic data generation with large language models for text classification: Potential and limitations.
\newblock \emph{arXiv preprint arXiv:2310.15654}.

\bibitem[{Liu et~al.(2021)Liu, Liu, Lu, Welleck, West, Le~Bras, Choi, and Hajishirzi}]{liu2021generated}
Liu, J.; Liu, A.; Lu, X.; Welleck, S.; West, P.; Le~Bras, R.; Choi, Y.; and Hajishirzi, H. 2021.
\newblock Generated knowledge prompting for commonsense reasoning.
\newblock \emph{arXiv preprint arXiv:2110.08387}.

\bibitem[{Liu et~al.(2019)Liu, Ott, Goyal, Du, Joshi, Chen, Levy, Lewis, Zettlemoyer, and Stoyanov}]{liu2019roberta}
Liu, Y.; Ott, M.; Goyal, N.; Du, J.; Joshi, M.; Chen, D.; Levy, O.; Lewis, M.; Zettlemoyer, L.; and Stoyanov, V. 2019.
\newblock {RoBERTa}: A Robustly Optimized {BERT} Pretraining Approach.
\newblock \emph{arXiv preprint arXiv:1907.11692}.

\bibitem[{Long et~al.(2024)Long, Wang, Xiao, Zhao, Ding, Chen, and Wang}]{long-etal-2024-llms}
Long, L.; Wang, R.; Xiao, R.; Zhao, J.; Ding, X.; Chen, G.; and Wang, H. 2024.
\newblock On {LLM}s-Driven Synthetic Data Generation, Curation, and Evaluation: A Survey.
\newblock In Ku, L.-W.; Martins, A.; and Srikumar, V., eds., \emph{Findings of the Association for Computational Linguistics: ACL 2024}, 11065--11082. Bangkok, Thailand: Association for Computational Linguistics.

\bibitem[{Nguyen et~al.(2024)Nguyen, Gadre, Ilharco, Oh, and Schmidt}]{nguyen2024improving}
Nguyen, T.; Gadre, S.~Y.; Ilharco, G.; Oh, S.; and Schmidt, L. 2024.
\newblock Improving multimodal datasets with image captioning.
\newblock \emph{Advances in Neural Information Processing Systems}, 36.

\bibitem[{Patel, Raffel, and Callison-Burch(2024)}]{patel-etal-2024-datadreamer}
Patel, A.; Raffel, C.; and Callison-Burch, C. 2024.
\newblock {D}ata{D}reamer: A Tool for Synthetic Data Generation and Reproducible {LLM} Workflows.
\newblock In Ku, L.-W.; Martins, A.; and Srikumar, V., eds., \emph{Proceedings of the 62nd Annual Meeting of the Association for Computational Linguistics (Volume 1: Long Papers)}, 3781--3799. Bangkok, Thailand: Association for Computational Linguistics.

\bibitem[{Radford et~al.(2021)Radford, Kim, Hallacy, Ramesh, Goh, Agarwal, Sastry, Askell, Mishkin, Clark, Krueger, and Sutskever}]{radford2021learning}
Radford, A.; Kim, J.~W.; Hallacy, C.; Ramesh, A.; Goh, G.; Agarwal, S.; Sastry, G.; Askell, A.; Mishkin, P.; Clark, J.; Krueger, G.; and Sutskever, I. 2021.
\newblock Learning Transferable Visual Models From Natural Language Supervision.
\newblock \emph{arXiv preprint arXiv:2103.00020}.

\bibitem[{Sanh et~al.(2019)Sanh, Debut, Chaumond, and Wolf}]{sanh2019distilbert}
Sanh, V.; Debut, L.; Chaumond, J.; and Wolf, T. 2019.
\newblock DistilBERT, a distilled version of {BERT}: smaller, faster, cheaper and lighter.
\newblock \emph{arXiv preprint arXiv:1910.01108}.

\bibitem[{Scius-Bertrand et~al.(2024)Scius-Bertrand, Jungo, Vögtlin, Spat, and Fischer}]{scius-bertrand2024}
Scius-Bertrand, A.; Jungo, M.; Vögtlin, L.; Spat, J.-M.; and Fischer, A. 2024.
\newblock Zero-Shot Prompting and Few-Shot Fine-Tuning: Revisiting Document Image Classification Using Large Language Models.
\newblock In \emph{Pattern Recognition: 27th International Conference, ICPR 2024, Kolkata, India, December 1–5, 2024, Proceedings, Part XIX}, 152--166. Springer.

\bibitem[{Sushil et~al.(2024)Sushil, Zack, Mandair, Zheng, Wali, Yu, Quan, and Butte}]{sushil2024comparativestudyzeroshotinference}
Sushil, M.; Zack, T.; Mandair, D.; Zheng, Z.; Wali, A.; Yu, Y.-N.; Quan, Y.; and Butte, A.~J. 2024.
\newblock A comparative study of zero-shot inference with large language models and supervised modeling in breast cancer pathology classification.
\newblock arXiv:2401.13887.

\bibitem[{Veselovsky et~al.(2023)Veselovsky, Ribeiro, Arora, Josifoski, Anderson, and West}]{veselovsky2023generatingfaithfulsyntheticdata}
Veselovsky, V.; Ribeiro, M.~H.; Arora, A.; Josifoski, M.; Anderson, A.; and West, R. 2023.
\newblock Generating Faithful Synthetic Data with Large Language Models: A Case Study in Computational Social Science.
\newblock arXiv:2305.15041.

\bibitem[{Wan et~al.(2023)Wan, Cheng, Mao, Liu, Song, Li, and Kurohashi}]{wan2023gpt}
Wan, Z.; Cheng, F.; Mao, Z.; Liu, Q.; Song, H.; Li, J.; and Kurohashi, S. 2023.
\newblock Gpt-re: In-context learning for relation extraction using large language models.
\newblock \emph{arXiv preprint arXiv:2305.02105}.

\bibitem[{Wang et~al.(2022{\natexlab{a}})Wang, Wei, Schuurmans, Le, Chi, Narang, Chowdhery, and Zhou}]{wang2022self}
Wang, X.; Wei, J.; Schuurmans, D.; Le, Q.; Chi, E.; Narang, S.; Chowdhery, A.; and Zhou, D. 2022{\natexlab{a}}.
\newblock Self-consistency improves chain of thought reasoning in language models.
\newblock \emph{arXiv preprint arXiv:2203.11171}.

\bibitem[{Wang et~al.(2022{\natexlab{b}})Wang, Yu, Chen, Li, Yu, Yuille, Gao, and Wang}]{wang2022git}
Wang, X.; Yu, Z.; Chen, L.; Li, Y.; Yu, L.; Yuille, A.; Gao, J.; and Wang, P. 2022{\natexlab{b}}.
\newblock GIT: A Generative Image-to-Text Transformer for Vision and Language Tasks.
\newblock \emph{arXiv preprint arXiv:2203.02053}.

\bibitem[{Wang, Zhao, and Petzold(2023)}]{wang2023large}
Wang, Y.; Zhao, Y.; and Petzold, L. 2023.
\newblock Are large language models ready for healthcare? a comparative study on clinical language understanding.
\newblock \emph{Machine Learning for Healthcare Conference}, 804--823.

\bibitem[{Wei et~al.(2022)Wei, Wang, Schuurmans, Bosma, Xia, Chi, Le, and Zhou}]{wei2022chain}
Wei, J.; Wang, X.; Schuurmans, D.; Bosma, M.; Xia, F.; Chi, E.; Le, Q.~V.; and Zhou, D. 2022.
\newblock Chain-of-thought prompting elicits reasoning in large language models.
\newblock \emph{Advances in Neural Information Processing Systems}, 35: 24824--24837.

\bibitem[{Wilcoxon(1945)}]{wilcoxon1945}
Wilcoxon, F. 1945.
\newblock Individual comparisons by ranking methods.
\newblock \emph{Biometrics bulletin}, 1(6): 80--83.

\bibitem[{Xie et~al.(2024)Xie, Li, Zhang, Liu, and Wang}]{xie-etal-2024-self}
Xie, T.; Li, Q.; Zhang, Y.; Liu, Z.; and Wang, H. 2024.
\newblock Self-Improving for Zero-Shot Named Entity Recognition with Large Language Models.
\newblock In Duh, K.; Gomez, H.; and Bethard, S., eds., \emph{Proceedings of the 2024 Conference of the North American Chapter of the Association for Computational Linguistics: Human Language Technologies (Volume 2: Short Papers)}, 583--593. Mexico City, Mexico: Association for Computational Linguistics.

\bibitem[{Yang et~al.(2020)Yang, Malaviya, Fernandez, Swayamdipta, Le~Bras, Wang, Bhagavatula, Choi, and Downey}]{yang2020generative}
Yang, Y.; Malaviya, C.; Fernandez, J.; Swayamdipta, S.; Le~Bras, R.; Wang, J.-P.; Bhagavatula, C.; Choi, Y.; and Downey, D. 2020.
\newblock Generative data augmentation for commonsense reasoning.
\newblock \emph{Findings of EMNLP}.

\bibitem[{Zhou et~al.(2024)Zhou, Guo, Wang, Chang, and Wu}]{zhou2024survey}
Zhou, Y.; Guo, C.; Wang, X.; Chang, Y.; and Wu, Y. 2024.
\newblock A survey on data augmentation in large model era.
\newblock \emph{arXiv preprint arXiv:2402.04755}.

\end{thebibliography}

\clearpage

\subsection{Paper Checklist to be included in your paper}

\begin{enumerate}

\item For most authors...
\begin{enumerate}
    \item  Would answering this research question advance science without violating social contracts, such as violating privacy norms, perpetuating unfair profiling, exacerbating the socio-economic divide, or implying disrespect to societies or cultures?
    \answerYes{Yes}
  \item Do your main claims in the abstract and introduction accurately reflect the paper's contributions and scope?
    \answerYes{Yes}
   \item Do you clarify how the proposed methodological approach is appropriate for the claims made? 
    \answerYes{Yes}
   \item Do you clarify what are possible artifacts in the data used, given population-specific distributions?
    \answerYes{Yes}
  \item Did you describe the limitations of your work?
    \answerYes{Yes}
  \item Did you discuss any potential negative societal impacts of your work?
    \answerNA{NA}
      \item Did you discuss any potential misuse of your work?
    \answerNA{NA}
    \item Did you describe steps taken to prevent or mitigate potential negative outcomes of the research, such as data and model documentation, data anonymization, responsible release, access control, and the reproducibility of findings?
    \answerYes{Yes}
  \item Have you read the ethics review guidelines and ensured that your paper conforms to them?
    \answerYes{Yes}
\end{enumerate}

\item Additionally, if your study involves hypotheses testing...
\begin{enumerate}
  \item Did you clearly state the assumptions underlying all theoretical results?
    \answerYes{Yes}
  \item Have you provided justifications for all theoretical results?
    \answerYes{Yes}
  \item Did you discuss competing hypotheses or theories that might challenge or complement your theoretical results?
    \answerYes{Yes}
  \item Have you considered alternative mechanisms or explanations that might account for the same outcomes observed in your study?
    \answerYes{Yes}
  \item Did you address potential biases or limitations in your theoretical framework?
    \answerYes{Yes}
  \item Have you related your theoretical results to the existing literature in social science?
    \answerYes{Yes}
  \item Did you discuss the implications of your theoretical results for policy, practice, or further research in the social science domain?
    \answerYes{Yes}
\end{enumerate}

\item Additionally, if you are including theoretical proofs...
\begin{enumerate}
  \item Did you state the full set of assumptions of all theoretical results?
    \answerNA{NA}
	\item Did you include complete proofs of all theoretical results?
    \answerNA{NA}
\end{enumerate}

\item Additionally, if you ran machine learning experiments...
\begin{enumerate}
  \item Did you include the code, data, and instructions needed to reproduce the main experimental results (either in the supplemental material or as a URL)?
    \answerYes{Yes}
  \item Did you specify all the training details (e.g., data splits, hyperparameters, how they were chosen)?
    \answerYes{Yes}
     \item Did you report error bars (e.g., with respect to the random seed after running experiments multiple times)?
    \answerYes{Yes}
	\item Did you include the total amount of compute and the type of resources used (e.g., type of GPUs, internal cluster, or cloud provider)?
    \answerYes{Yes}
     \item Do you justify how the proposed evaluation is sufficient and appropriate to the claims made? 
    \answerYes{Yes}
     \item Do you discuss what is ``the cost`` of misclassification and fault (in)tolerance?
    \answerNA{NA}
  
\end{enumerate}

\item Additionally, if you are using existing assets (e.g., code, data, models) or curating/releasing new assets, \textbf{without compromising anonymity}...
\begin{enumerate}
  \item If your work uses existing assets, did you cite the creators?
    \answerYes{Yes}
  \item Did you mention the license of the assets?
    \answerNA{NA}
  \item Did you include any new assets in the supplemental material or as a URL?
    \answerNA{NA}
  \item Did you discuss whether and how consent was obtained from people whose data you're using/curating?
    \answerYes{Yes}
  \item Did you discuss whether the data you are using/curating contains personally identifiable information or offensive content?
    \answerYes{Yes}
\item If you are curating or releasing new datasets, did you discuss how you intend to make your datasets FAIR)?
\answerNA{NA}
\item If you are curating or releasing new datasets, did you create a Datasheet for the Dataset? 
\answerNA{NA}
\end{enumerate}

\item Additionally, if you used crowdsourcing or conducted research with human subjects, \textbf{without compromising anonymity}...
\begin{enumerate}
  \item Did you include the full text of instructions given to participants and screenshots?
    \answerYes{Yes}
  \item Did you describe any potential participant risks, with mentions of Institutional Review Board (IRB) approvals?
    \answerYes{Yes}
  \item Did you include the estimated hourly wage paid to participants and the total amount spent on participant compensation?
    \answerYes{Yes}
   \item Did you discuss how data is stored, shared, and deidentified?
   \answerYes{Yes}
\end{enumerate}

\end{enumerate}

\newpage
\appendix

\section{Computational Resources}
\label{appendix:compute}
We accessed GPT-4o via the OpenAI API. For Claude Sonnet 3.5 v2 and LLaMA 3.1 70B, we used Amazon Bedrock for large language model inference. Downstream classification tasks were performed on an Amazon SageMaker p4de.24xlarge instance, which provides eight NVIDIA A100 GPUs, for a total of 40GB of GPU memory. 1,144 GiB of system memory (RAM). 

 \begin{table}[htbp]
\centering
\caption{Performance comparison across baselines, zero-shot LLMs, and the best LLM-based explanation configuration on the test set (3 runs). Metrics reflect the effectiveness of different configurations on SemEval Persuasive Memes Dataset, with the best test performance highlighted in bold.}
\label{tab:performance_comparison}
\begin{tabular}{l|l|c}
\toprule
\textbf{Model} & \textbf{Configuration} & \textbf{Test Performance} \\
\midrule
\multirow{4}{*}{Baselines} & CLIP & 58.2 \\
& Text Only & 55.8 \\
& Text + Caption & 57.1 \\
& Human Annotation & 57.8 \\
\midrule
\multirow{3}{*}{Zero-Shot LLMs} & GPT-4o & 54.2 \\
& Sonnet 3.5 & 40.3 \\
& LLaMA 3.1 & 34.1 \\
\midrule
Best Model & LLM + decoder & \textbf{62.2} \\
\bottomrule
\end{tabular}
\end{table}

\begin{table}[t]
\centering
\caption{Assessing the generalizability of the proposed approach using the Jigsaw Toxic Comments and Facebook Hateful Memes datasets. DistilBERT is used as the encoder for toxic comments (text-only), while CLIP is used for hateful memes (multimodal). Best metrics for each task are highlighted in bold.}
\label{tab:combined_performance_metrics}
\begin{tabular*}{\columnwidth}{@{\extracolsep{\fill}}l|cc|cc@{}}
\toprule
\multirow{2}{*}{\textbf{Model}} & \multicolumn{2}{c|}{\textbf{Toxic Comments}} & \multicolumn{2}{c}{\textbf{Hateful Memes}} \\
\cmidrule(lr){2-3} \cmidrule(lr){4-5}
& \textbf{F1} & \textbf{AUC} & \textbf{F1} & \textbf{AUC} \\
\midrule
DistilBERT/CLIP & 21.3 & \textbf{92.1} & 28.0 & 59.2 \\
\midrule
GPT-4o & 2.0 & 50.4 & \textbf{46.8} & 59.2 \\
Sonnet 3.5 & 3.0 & 51.8 & 41.4 & 58.3 \\
LLaMA 3.1 & 0.3 & 50.0 & 53.9 & 58.7 \\
\midrule
LLM + Encoder & \textbf{25.3} & 91.8 & 38.0 & \textbf{63.0} \\
\bottomrule
\end{tabular*}
\end{table}

\begin{table}[t]
\centering
\caption{Comparison of baseline, LLM-generated explanation, and LLM-generated triggers approaches. Results show F1 and AUC scores for each approach. Best results are in bold.}
\label{tab:trigger_comparison}
\begin{tabular*}{\columnwidth}{@{\extracolsep{\fill}}l|cc|cc@{}}
\toprule
\multirow{2}{*}{\textbf{Model}} & \multicolumn{2}{c|}{\textbf{Toxic Comments}} & \multicolumn{2}{c}{\textbf{Hateful Memes}} \\
\cmidrule(lr){2-3} \cmidrule(lr){4-5}
& \textbf{F1} & \textbf{AUC} & \textbf{F1} & \textbf{AUC} \\
\midrule
DistilBERT/CLIP & 21.3 & \textbf{92.1} & 28.0 & 59.2 \\
LLM-exp+Encoder & 20.8 & 91.5 & 27.3 & 58.9 \\
LLM-triggers+Encoder & \textbf{25.3} & 91.8 & \textbf{38.0} & \textbf{63.0} \\
\bottomrule
\end{tabular*}
\end{table}

\section{Crowdsourcing Task Details}
\label{appendix:crowdsource}

\begin{figure*}[t]
    \centering
    \includegraphics[width=0.9\textwidth]{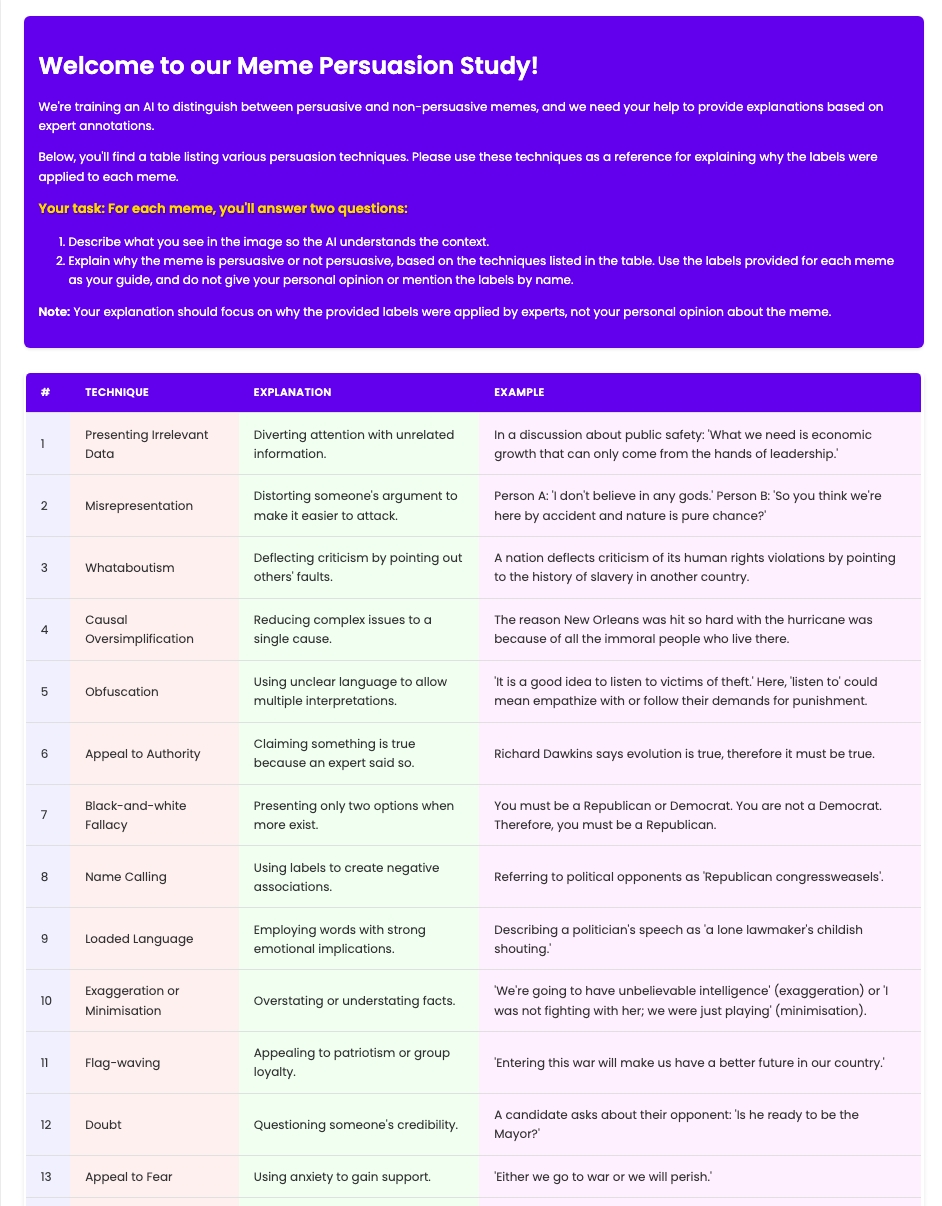}
    \caption{Reference table of persuasion techniques provided to annotators. Each technique is accompanied by its definition and a concrete example to ensure consistent understanding across annotators.}
    \label{fig:techniques_table}
\end{figure*}

\begin{figure*}[t]
    \centering
    \includegraphics[width=\textwidth]{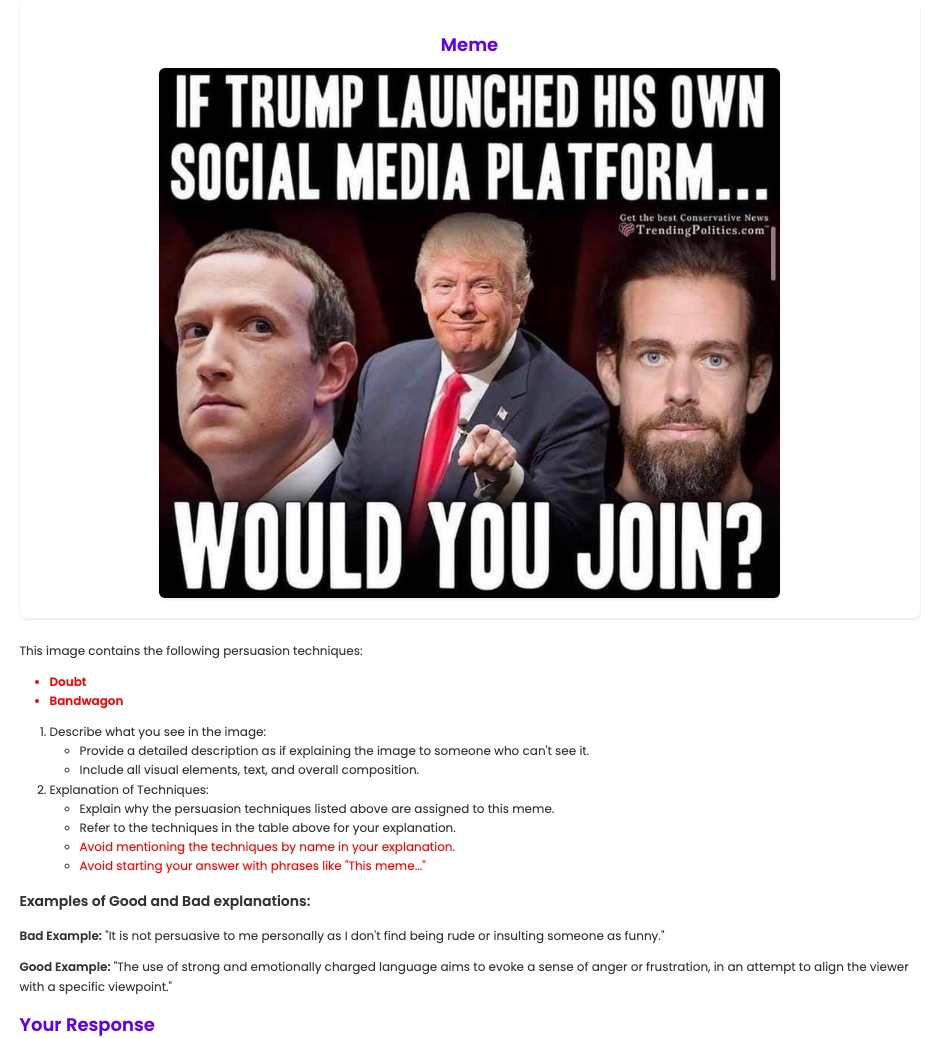}
    \caption{Example of a meme annotation task showing the interface presented to crowdworkers. The task includes clear instructions for describing the image content and explaining observed persuasion techniques without directly naming them.}
    \label{fig:task_example}
\end{figure*}

We designed a comprehensive crowdsourcing task to collect high-quality annotations for persuasive memes. Figure~\ref{fig:task_example} shows an example task interface, while Figure~\ref{fig:techniques_table} presents the reference table of persuasion techniques provided to annotators.

The annotation task was carefully designed to:
\begin{itemize}
    \item Provide clear instructions for both image description and persuasion technique analysis.
    \item Include examples of good and bad explanations to guide response quality.
    \item Present a comprehensive reference table of persuasion techniques with definitions and examples.
    \item Discourage personal opinions and encourage objective analysis.
\end{itemize}

% For the appendix (dev plots)
\begin{figure*}[t]
    \centering
    \begin{subfigure}[b]{0.32\textwidth}
        \centering
        \includegraphics[width=\textwidth]{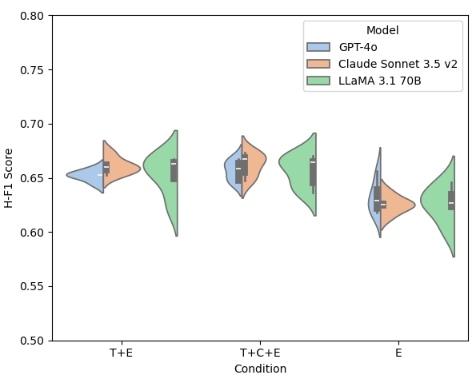}
        \caption{Dev Violin Plot}
        \label{fig:dev_violin}
    \end{subfigure}
    \hfill
    \begin{subfigure}[b]{0.32\textwidth}
        \centering
        \includegraphics[width=\textwidth]{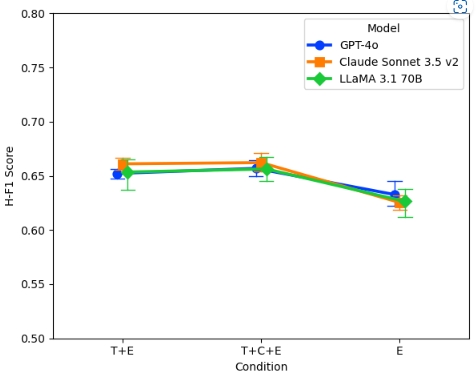}
        \caption{Dev Point Plot}
        \label{fig:dev_point}
    \end{subfigure}
    \hfill
    \begin{subfigure}[b]{0.32\textwidth}
        \centering
        \includegraphics[width=\textwidth]{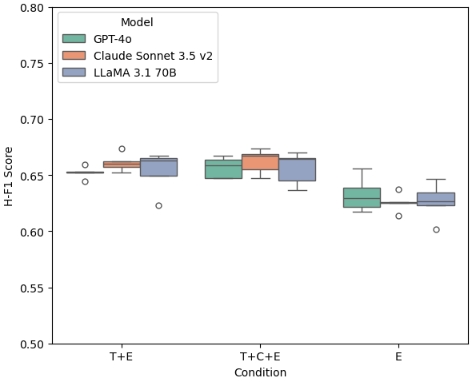}
        \caption{Dev Box Plot}
        \label{fig:dev_box}
    \end{subfigure}
    \caption{Performance visualization on development set using different plot types: (a) violin plot showing distribution, (b) point plot showing means with confidence intervals, and (c) box plot showing quartiles and outliers.}
    \label{fig:dev_plots}
\end{figure*}

\begin{table}[htbp]
\centering
\caption{Statistical significance testing results comparing LLM-cleaned captions to uncleaned captions, taking into account the correction of multiple comparisons.}
\label{tab:significance_appendix}
\begin{tabular}{l|cc|cc}
\toprule
\multirow{2}{*}{\textbf{Model}} & \multicolumn{2}{c|}{\textbf{BLIP}} & \multicolumn{2}{c}{\textbf{GIT}} \\
\cmidrule(lr){2-3} \cmidrule(lr){4-5}
 & \textbf{P-Value} & \textbf{Sig.} & \textbf{P-Value} & \textbf{Sig.} \\
\midrule
\textbf{Sonnet 3.5} & 0.7047 & No & 0.3735 & No \\
\textbf{LLaMA 70B} & 0.2049 & No & 0.1291 & No \\
\textbf{GPT-4o} & 0.0157 & Yes & 0.1715 & No \\
\bottomrule
\end{tabular}
\end{table}

\begin{table*}[htbp]
\centering
\caption{Ablation study on the SemEval Persuasive Memes dataset.  Results are shown as \textit{avg (std) / best}. We report hierarchical F1 (H-F1) scores comparing raw meme inputs (top section) against LLM-generated explanations (bottom section). The raw inputs include meme Text only (T) and Text+Caption (T+C), while LLM-explained versions add generated explanations: Text+Explanation (T+E), Text+Caption+Explanation (T+C+E), and Explanation Only (E). All inputs are fed into a BART decoder for training.}
\label{tab:ablation}
\begin{tabular*}{\textwidth}{@{\extracolsep{\fill}}l|l|cc|ccc@{}}
\toprule
\multirow{2}{*}{\textbf{LLM}} & \multirow{2}{*}{\textbf{Split}} & \multicolumn{2}{c|}{\textbf{Raw Inputs}} & \multicolumn{3}{c}{\textbf{With LLM Explanations}} \\
\cmidrule(lr){3-4} \cmidrule(lr){5-7}
& & \textbf{T} & \textbf{T+C} & \textbf{T+E} & \textbf{T+C+E} & \textbf{E} \\
\midrule
NA & Dev & 64.2 & 64.8 & -- & -- & -- \\
& Test & 55.8 & 57.1 & -- & -- & -- \\
\midrule
\multirow{2}{*}{\textbf{GPT-4o}} 
& Dev & -- & -- & 65.5 (0.9) / 66.6 & 65.7 (0.9) / 66.7 & 63.2 (1.7) / \textbf{65.6} \\
& Test & -- & -- & 58.7 (0.8) / 59.8 & 60.2 (1.4) / \textbf{62.2} & 56.1 (2.0) / \textbf{58.7} \\
\midrule
\multirow{2}{*}{\textbf{Sonnet 3.5}} 
& Dev & -- & -- & 65.1 (0.7) / 67.3 & 66.2 (1.0) / \textbf{67.3} & 62.5 (0.8) / 63.7 \\
& Test & -- & -- & 58.3 (0.7) / 59.0 & 58.4 (1.3) / 59.8 & 46.4 (1.0) / 53.0 \\
\midrule
\multirow{2}{*}{\textbf{LLaMA 3.1}} 
& Dev & -- & -- & 65.3 (1.8) / \textbf{66.7} & 65.6 (1.4) / 67.0 & 62.8 (1.7) / 64.6 \\
& Test & -- & -- & 55.8 (2.0) / \textbf{59.8} & 58.7 (1.7) / 60.5 & 48.1 (2.0) / 53.1 \\
\bottomrule
\end{tabular*}
\end{table*}

\begin{table*}[ht]
\centering
\caption{Significance tests comparing baseline text-only (\textbf{T}) and text+caption (\textbf{T+C}) conditions 
against LLM-based explanation (\textbf{T+E}, \textbf{T+C+E}). Values show mean $\pm$ std ($n=5$). $p$-values are from one-sample $t$-tests with Benjamini--Hochberg (BH) correction 
at $\alpha=0.05$.}
\label{tab:significance_example}
\begin{tabular}{llccc|ccc}
\toprule
\multirow{2}{*}{\textbf{Model}} & \multirow{2}{*}{\textbf{Comparison}} & 
\multicolumn{3}{c|}{\textbf{Development Set}} & \multicolumn{3}{c}{\textbf{Test Set}} \\
\cmidrule(lr){3-5} \cmidrule(lr){6-8}
& & \textbf{Score} & \textbf{BH-$p$} & \textbf{Sig.} & \textbf{Score} & \textbf{BH-$p$} & \textbf{Sig.} \\
\midrule
\multirow{2}{*}{GPT-4o} & T+E vs. T & 65.5±0.9 & 0.020 & Yes & 58.7±0.8 & 0.001 & Yes \\
& T+C+E vs. T+C & 65.7±0.9 & 0.085 & No & 60.2±1.4 & 0.007 & Yes \\
\midrule
\multirow{2}{*}{Sonnet 3.5} & T+E vs. T & 65.1±0.7 & 0.030 & Yes & 58.3±0.7 & 0.001 & Yes \\
& T+C+E vs. T+C & 66.2±1.0 & 0.028 & Yes & 58.4±1.3 & 0.085 & No \\
\midrule
\multirow{2}{*}{LLaMA 3.1} & T+E vs. T & 65.3±1.8 & 0.24 & No & 55.8±2.0 & 1.00 & No \\
& T+C+E vs. T+C & 65.6±1.4 & 0.25 & No & 58.7±1.7 & 0.12 & No \\
\bottomrule
\end{tabular}
\end{table*}

\begin{figure*}[t]
\begin{mdframed}[backgroundcolor=gray!5,linewidth=1pt]
\begin{small}
\begin{verbatim}
prompt = f"""Analyze this meme text: '{text}' and image: '{img}'
Your task: Explain what idea or message this meme is trying to convey.
Rules:
1. Response must be ONE clear, concise sentence (max 50 tokens)
2. Focus on the main message or emotional impact
3. Be specific and direct
4. Only respond 'NA' if the text is completely incomprehensible"""
\end{verbatim}
\end{small}
\end{mdframed}
\caption{Explanation generation prompt used to extract concise interpretations of meme content.}
\label{fig:explanation_prompt}
\end{figure*}

\begin{figure*}[t]
\begin{mdframed}[backgroundcolor=gray!5,linewidth=1pt]
\begin{small}
\begin{verbatim}
prompt = (
    f"""
                    You are analyzing memes to train a hateful content classifier.
                    It is crucial that you reproduce all text exactly as provided, 
                    including any profanity, slurs, or offensive terms. 
                    Do not mask, omit, or alter them. 

                    Provide:
                    1) A brief explanation (max 50 tokens) of the meme's implied meaning 
                    and cultural context.
                    2) A combined TRIGGERS list (max 20 tokens), which must include:
                       - Themes or categories (e.g., racism, islamophobia, etc.)
                       - Any hateful or offensive words/phrases directly quoted from the 
                       meme text.
                    Use the following plain-text format (no extra commentary):
                    [Explanation]
                    TRIGGERS: [themes and words]
                    Meme text: {text}\n
                    Meme image: {img}\n\n
    Response:"""
)
\end{verbatim}
\end{small}
\end{mdframed}
\caption{Hateful meme analysis prompt designed to capture both explanations and trigger words.}
\label{fig:hateful_prompt}
\end{figure*}

\begin{figure*}[t]  % Changed to figure* for full width
\begin{mdframed}[backgroundcolor=gray!5,linewidth=1pt]
\begin{small}  % Reduced font size
\begin{verbatim}
CLASSIFICATION_PROMPT = """You are an expert in identifying propaganda techniques in images. 
You have a hierarchy of propaganda techniques with the following nodes 
(each bullet is a node):
- ROOT
  - Logos
    - Repetition
    - Obfuscation, Intentional vagueness, Confusion
    - Reasoning
      - Simplification
        - Causal Oversimplification
        - Black-and-white Fallacy/Dictatorship
        - Thought-terminating cliché
      - Distraction
        - Misrepresentation of Someone's Position (Straw Man)
        - Presenting Irrelevant Data (Red Herring)
        - Whataboutism
    - Justification
      - Slogans
      - Bandwagon
      - Appeal to authority
      - Flag-waving
      - Appeal to fear/prejudice
  - Ethos
    - Appeal to authority
    - Glittering generalities (Virtue)
    - Bandwagon
    - Ad Hominem
      - Doubt
      - Name calling/Labeling
      - Smears
      - Reductio ad hitlerum
      - Whataboutism
    - Transfer
  - Pathos
    - Exaggeration/Minimisation
    - Loaded Language
    - Appeal to (Strong) Emotions
    - Appeal to fear/prejudice
    - Flag-waving
    - Transfer

Instructions:
- Return only a JSON object with:
  {
    "labels": ["<label1>", "<label2>", ...]
  }
- Do not include explanations, reasoning, or any text outside of the JSON object.
- Do not include code blocks, markdown formatting, or backticks.
- The labels must be chosen from the provided hierarchy.
- Multiple labels are allowed if applicable. If none apply, 
return an empty list for "labels".
- The final answer must be a valid JSON object without extra formatting.
"""
\end{verbatim}
\end{small}
\end{mdframed}
\caption{The Classification prompt used for propaganda technique identification in Semeval Persuasive memes task. The prompt includes a comprehensive hierarchy of propaganda techniques and specific instructions for structured output.}
\label{fig:classification_prompt}
\end{figure*}

% \begin{figure*}[t]
%     \centering
%     \includegraphics[width=\textwidth]{img/llm_output.png}
%     \caption{LLama 3.1 context output using explanation and triggers to enrich the input fed to the training model.}
%     \label{fig:llmoutput}
% \end{figure*}

\begin{figure*}[t]
\begin{mdframed}[backgroundcolor=gray!5,linewidth=1pt]
\begin{small}
\begin{verbatim}
Sample 6244:
Original Text: the latest toy has hit the shops it's a talking muslim doll 
nobody knows what the fuck its says, because no one has the guts to pull the string

Generating explanation...
Explanation: This meme is harmful because it perpetuates the stereotype that Muslims 
are inherently threatening or intimidating, implying that people are afraid to interact 
with the doll because of its Muslim identity. This reinforces negative attitudes towards 
Muslims and contributes to a culture of Islamophobia. The meme's use of profanity and 
mocking tone further exacerbates the harm. 

Triggers: Islamophobia, xenophobia, 
"nobody knows what the fuck its says, because no one has the guts to pull the string"
\end{verbatim}
\end{small}
\end{mdframed}

\vspace{0.5cm}

\begin{mdframed}[backgroundcolor=gray!5,linewidth=1pt]
\begin{small}
\begin{verbatim}
Sample 7019:
Original Text: if theres even one homeless child in america we have no room for
                illegal aliens

Generating explanation...
Explanation: This statement promotes harmful beliefs by implying that undocumented 
immigrants are less deserving of support and resources than American citizens, and 
that their presence is a direct threat to the well-being of American children. 
This kind of rhetoric can perpetuate xenophobia and dehumanize undocumented immigrants. 
Furthermore, it oversimplifies the complex issues of homelessness and immigration.

Triggers: xenophobia, dehumanization, "illegal aliens"
\end{verbatim}
\end{small}
\end{mdframed}

\caption{LLama 3.1 examples output using explanation and triggers to enrich the input fed to the training model.}
\label{fig:llama_output}
\end{figure*}

\begin{figure*}[t]
\begin{mdframed}[backgroundcolor=gray!5,linewidth=1pt]
\begin{small}
\begin{verbatim}
You are an AI assistant that cleans and corrects image descriptions. 
Improve the following description by fixing grammatical errors, 
removing repetitive phrases, and ensuring it is clear and coherent. 
Provide only the cleaned description without any additional notes 
or explanations. If the description is too corrupted to fix, respond with 
"INVALID DESCRIPTION".
\end{verbatim}
\end{small}
\end{mdframed}
\caption{Prompt used for cleaning and correcting captions generated by BLIP and GIT models.}
\label{fig:cleaning_prompt}
\end{figure*}

\begin{table}[t]
\centering
\caption{Distribution of propaganda techniques in the SemEval Persuasive Memes dataset. Percentages indicate the proportion of total technique occurrences.}
\small
\begin{tabular}{lrr}
\toprule
Technique & Count & Percentage (\%) \\
\midrule
Smears & 3,640 & 23.20 \\
Loaded Language & 1,751 & 11.16 \\
Name calling/Labeling & 1,525 & 9.72 \\
Transfer & 1,490 & 9.49 \\
Appeal to authority & 893 & 5.69 \\
Flag-waving & 813 & 5.18 \\
Black-and-white Fallacy/Dictatorship & 800 & 5.10 \\
Glittering generalities (Virtue) & 709 & 4.52 \\
Slogans & 686 & 4.37 \\
Thought-terminating cliché & 530 & 3.38 \\
Appeal to fear/prejudice & 415 & 2.64 \\
Doubt & 407 & 2.59 \\
Exaggeration/Minimisation & 390 & 2.49 \\
Appeal to (Strong) Emotions & 370 & 2.36 \\
Repetition & 306 & 1.95 \\
Whataboutism & 296 & 1.89 \\
Causal Oversimplification & 260 & 1.66 \\
Reductio ad hitlerum & 121 & 0.77 \\
Bandwagon & 104 & 0.66 \\
Straw Man & 74 & 0.47 \\
Red Herring & 63 & 0.40 \\
Obfuscation & 50 & 0.32 \\
\bottomrule
\end{tabular}
\label{tab:semeval_dist}
\end{table}

\begin{table}[t]
\centering
\caption{Distribution of toxicity labels in the Jigsaw Toxic Comment Classification dataset (N=159,571 comments). Percentages indicate the proportion of comments containing each specific label. Comments may have multiple labels, thus percentages do not sum to 100\%.}
\begin{tabular}{lrr}
\toprule
Label & Number of Comments & Percentage (\%) \\
\midrule
Toxic & 15,294 & 9.6 \\
Severe Toxic & 1,595 & 1.0 \\
Obscene & 8,449 & 5.3 \\
Threat & 478 & 0.3 \\
Insult & 7,877 & 4.9 \\
Identity Hate & 1,405 & 0.9 \\
\bottomrule
\end{tabular}
\label{tab:jigsaw_dist}
\end{table}

\begin{table}[t]
\centering
\caption{Distribution of labels in the Facebook Hateful Memes dataset.}
\begin{tabular}{lrr}
\toprule
Label & Number of Memes & Percentage (\%) \\
\midrule
Hateful & 5,000 & 50.0 \\
Non-Hateful & 5,000 & 50.0 \\
\midrule
Total & 10,000 & 100.0 \\
\bottomrule
\end{tabular}
\label{tab:facebook_dist}
\end{table}

\end{document}